\documentclass[11pt]{article}

\usepackage[final]{acl}
\usepackage{pdfpages}
\usepackage{times}
\usepackage{latexsym}
\usepackage[T1]{fontenc}
\usepackage[utf8]{inputenc}
\usepackage{microtype}
\usepackage{inconsolata}
\usepackage{graphicx}

\usepackage{amsmath,amssymb,amsfonts}
\usepackage{booktabs}
\usepackage{array}
\usepackage{xspace}
\usepackage{tabu}
\usepackage{multirow}
\usepackage{enumitem}
\usepackage{subcaption}
\usepackage[most]{tcolorbox}
\usepackage{listings}
\usepackage{xcolor}
\usepackage{cuted}

\lstdefinestyle{promptstyle}{
    basicstyle=\ttfamily\footnotesize,
    breaklines=true,
    breakatwhitespace=false,
    columns=fullflexible,
    keepspaces=true,
    frame=none,
    backgroundcolor=\color{white},
    aboveskip=0pt,
    belowskip=0pt,
    escapechar=@
}

\newtcolorbox{promptbox}[1][]{
    enhanced,
    colback=gray!5,
    colframe=black!70,
    boxrule=0.5pt,
    arc=2pt,
    left=4pt, right=4pt, top=3pt, bottom=2pt,
    fonttitle=\bfseries,
    title={System Prompt},
    breakable,
    #1
}

\newcommand{\dname}{\textsc{M$^2$LongBench}\xspace}
\newcommand{\mname}{\textsc{Deep-Reporter}\xspace}

\usepackage{pifont}
\usepackage{tikz}
\definecolor{darkgreen}{RGB}{50,100,0}
\definecolor{darkred}{RGB}{200, 0, 0}
\definecolor{citecolor}{HTML}{0071BC}
\definecolor{linkcolor}{HTML}{ED1C24}
\newcommand{\cmark}{\textcolor{darkgreen}{\ding{51}}}
\newcommand{\xmark}{\textcolor{darkred}{\ding{55}}}
\newcommand{\cxmark}{%
    \tikz[baseline=(char.base)]{
        \node(char)[shape=rectangle, inner sep=0] {\textcolor{darkgreen}{\ding{51}}};
        \draw[darkred, thick, scale=0.16] (0.3,-0.2) -- (-0.2,0.3);
    }
}

\usepackage{hyperref}
\hypersetup{
    colorlinks=true,
    linkcolor=linkcolor,
    citecolor=citecolor
}

\title{%
\begin{minipage}{0.8\textwidth}
\centering
\raisebox{-0.35\height}[0pt][0pt]{%
  \includegraphics[height=1.9\baselineskip]{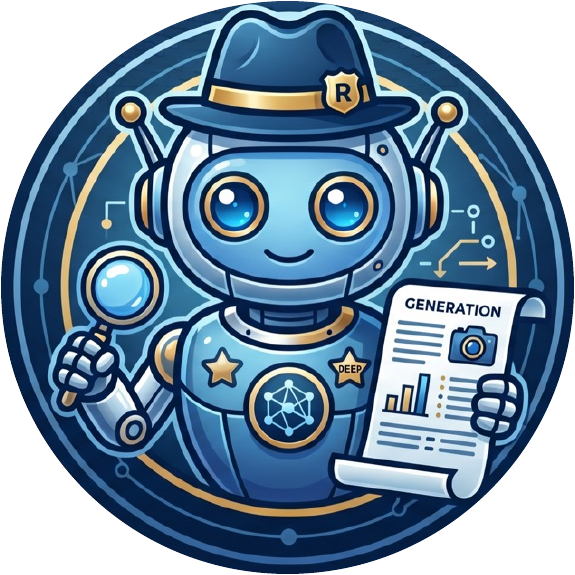}}%
\hspace{-8pt}%
\begin{minipage}[c]{0.85\textwidth}
\centering
Deep-Reporter: Deep Research for Grounded \\
Multimodal Long-Form Generation
\end{minipage}%
\end{minipage}%
}

\author{Fangda Ye$^1$$^{\dagger}$, Zhifei Xie$^2$$^{\dagger}$, Yuxin Hu$^1$$^{\dagger}$, Yihang Yin$^3$,    \\
        {\bf Shurui Huang$^1$, Shikai Dong$^4$, Jianzhu Bao$^2$, Shuicheng Yan$^1$} \\
        $^1$ National University of Singapore \quad
        $^2$ Nanyang Technological University \\
        $^3$ University of Edinburgh \quad
        $^4$ Beijing Institute of Technology}

\begin{document}
\maketitle

\def\thefootnote{$\dagger$}\footnotetext{These authors contributed equally. Correspondence to: Fangda Ye \texttt{<fangda.ye@u.nus.edu>} and Shuicheng Yan \texttt{<yansc@nus.edu.sg>}.}
\def\thefootnote{\arabic{footnote}}

\begin{abstract}

Recent agentic search frameworks enable deep research via iterative planning and retrieval, reducing hallucinations and enhancing factual grounding.
However, they remain text-centric, overlooking the multimodal evidence that characterizes real-world expert reports.
We introduce a pressing task: multimodal long-form generation. Accordingly, we propose \mname, a unified agentic framework for grounded multimodal long-form generation. It orchestrates: (i) \textit{Agentic Multimodal Search and Filtering} to retrieve and filter textual passages and information-dense visuals; (ii) \textit{Checklist-Guided Incremental Synthesis} to ensure coherent image-text integration and optimal citation placement; and (iii) \textit{Recurrent Context Management} to balance long-range coherence with local fluency. We develop a rigorous curation pipeline producing 8K high-quality agentic traces for model optimization.
We further introduce \dname, a comprehensive testbed comprising 247 research tasks across 9 domains and a stable multimodal sandbox. It enables unified multimodal assessment, fair comparison, and accessible evaluation without commercial APIs.
Extensive experiments demonstrate that long-form multimodal generation is a challenging task, especially in multimodal selection and integration, and effective post-training can bridge the gap.
Our code is available at \url{https://github.com/fangda-ye/Deep-Report}.

\end{abstract}

\section{Introduction}
\label{sec:introduction}

\begin{figure*}[t]
    \centering
    \includegraphics[width=0.95\linewidth]{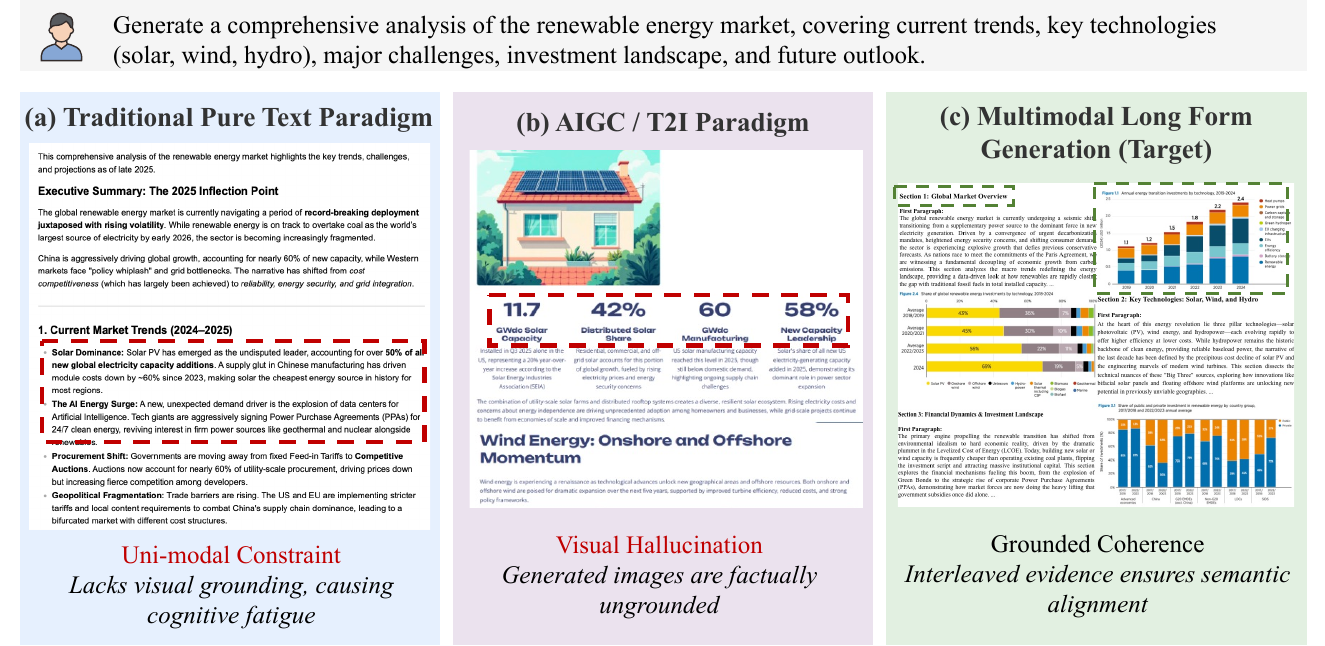}
    \caption{\textbf{Comparison of paradigms in long-form report generation.} While traditional systems (a) lack visual engagement and pure T2I approaches (b) suffer from factual hallucinations and narrative fragmentation, our method (c) achieves high coherence and factuality by retrieving and integrating real-world visual evidence.}
    \vspace{-1.5em}
    \label{fig:task}
\end{figure*}

% The advent of LLMs has driven remarkable progress in traditional question answering
LLMs have driven remarkable progress in question answering~\cite{rajpurkar2016squad,kwiatkowski2019natural,rein2024gpqa}. Building upon this foundation, recent research has increasingly focused on complex long-form generation, including Wiki article~\cite{shao2024assisting,yang2025wikiautogen}, academic paper~\cite{lu2024ai,ghafarollahi2025sciagents}, and industry reports~\cite{tian2025template}. These systems typically tackle multifaceted queries, such as "\textit{Generate an analysis of the renewable energy market}", which demand extensive factual grounding across multiple sources. Existing approaches employ agentic search~\citep{zhang2025reasonrag,jin2025search,li2025reinforcement,team2026mirothinker}, where autonomous agents iteratively plan queries, retrieve information from external sources, and synthesize findings. This paradigm mitigates hallucinations and enhances answer factuality.

However, existing long-form generation systems remain text-centric (Figure~\ref{fig:task}a), overlooking multimodal generation~\cite{xi2025survey,gu2025rapid,bai2024longwriter,bao2022aeg,ye2026miroeval}. While recent works~\cite{yang2025multimodal,xia2024mmed,zhang2026abtest} reveal that real-world expert reports rely heavily on visual evidence including: tables, charts, infographics, complex layouts, etc.
Incorporating such multimodal elements fundamentally enhances both interpretability and engagement of generated reports. Yet this integration introduces three critical challenges:
(1) \textbf{\textit{Agentic Multimodal Retrieval}}:
How can we effectively adapt text-based agentic search frameworks to retrieve, filter, and integrate multimodal information?
(2) \textbf{\textit{Coherent Multimodal Long-Form Generation}}: How can we maintain \textit{global sectional coherence}, achieve \textit{image-text consistency}, and manage extensive \textit{multimodal context}?
(3) \textbf{\textit{Evaluation of Multimodal Generation}}:
Given open-ended nature of agentic search and long-form generation, how can we evaluate in a \textit{rigorous and reproducible setting}?

To address these challenges, we introduce \mname ($\S$\ref{sec:method}), a unified agentic framework that adapts existing agentic search and long-form generation paradigms to multimodal setting.
\mname supports comprehensive multimodal agentic search, achieved by collaborative interactions to decompose complex queries, iteratively retrieves both textual and visual passages, refine/filter relevant evidence.
Subsequently, \mname employs a "Checklist-Guided Incremental Synthesis" mechanism, formalizing the report structure into "Semantic Anchors" and utilizing recurrent context update strategy, to ensure coherent interleaving of text and images throughout the report (as shown in Figure~\ref{fig:task}c).
To equip open-source models with these capabilities, we propose a pipeline to curate high-quality agentic trace ($\S$\ref{subsec:training}). We specifically target two critical competencies: (1) \textbf{\textit{Agentic Multimodal Search}}: teaching agents to strategically plan queries, retrieve heterogeneous content, and filter high-value visual evidence, and (2) \textbf{\textit{Coherent Multimodal Synthesis}}: demonstrating when and where to cite visual elements while maintaining global coherence via checklist-driven generation.
Through rigorous post-processing with LLM-based scoring and human verification on visual hallucinations, we curate 8k expert trajectories to enhance model multimodal competencies.

\begin{figure*}[t]
    \centering
    \includegraphics[width=1\linewidth]{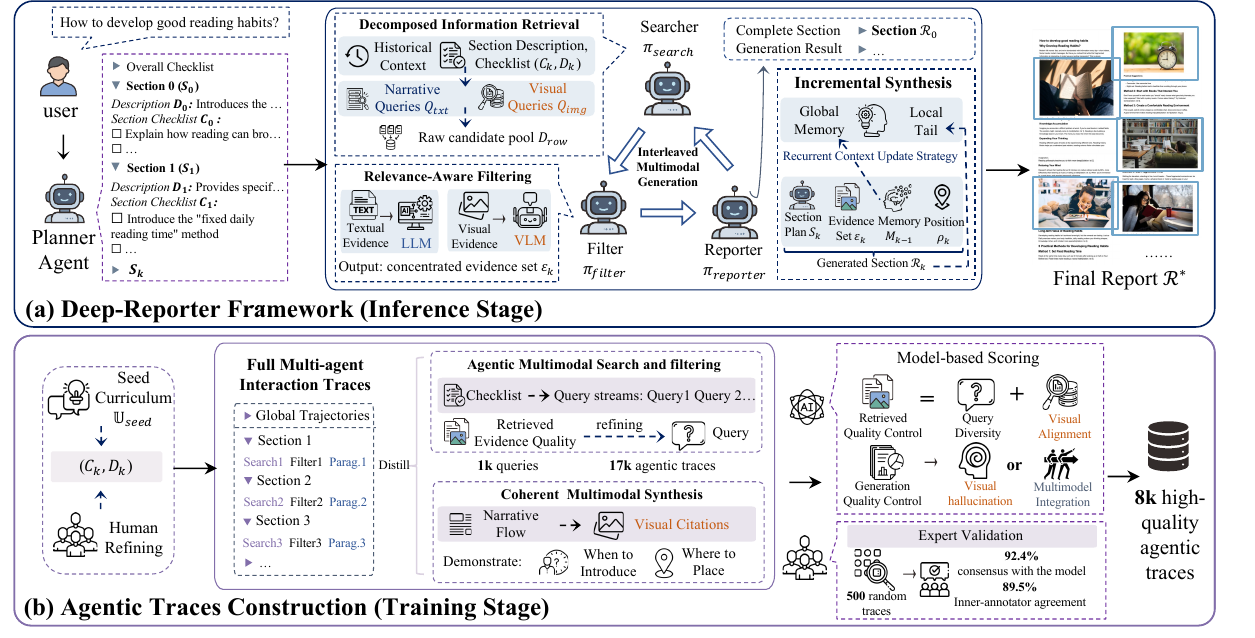}
    \vspace{-2.0em}
    \caption{\textbf{\mname Architecture.} (a) The multi-agent framework orchestrates planning, multimodal information seeking, and incremental writing to generate professional reports. (b) The data synthesis pipeline constructs high-quality expert trajectories to equip open-weight models with deep research capabilities.}
    \vspace{-1.0em}
    \label{fig:deep_reporter}
\end{figure*}

\vspace{-0.8em}
To address the lack of standardized multimodal testbeds, we construct \dname ($\S$\ref{sec:benchmark}), a benchmark comprising 247 complex research tasks across 9 domains. Unlike existing benchmarks~\cite{wu2025longeval,du2025deepresearch} that are either text-only or rely on dynamic web APIs, \dname establishes a stable multimodal sandbox containing 95K images and 108M text chunks (aggregated from 45K webpages and 6.4K PDFs), with annotations averaging 102 images and 168 text chunks as ground-truth evidence per report. This enable rigorous evaluation of visual retrieval and image-text coherence in long-form generation. \dname addresses three critical gaps:
(1) \textit{Unified Multimodal Assessment}: simultaneous evaluation of textual and visual retrieval precision, along with image-text integration quality in generated reports;
(2) \textit{Fairness \& Transparency}: researchers can isolate retrieval algorithm contributions within a controlled corpus~\cite{chen2025browsecompplus}, eliminating temporal drift from evolving web content and opacity of black-box APIs;
(3) \textit{Accessibility}: the self-contained sandbox removes dependencies on expensive commercial search APIs, lowering barriers while maintaining evaluation rigor.
Experiments on \dname confirm that \mname substantially outperforms RAG baselines, and our curated agentic traces significantly boost multimodal evidence selection and generation quality.
In summary, we present a holistic solution for grounded multimodal long-form generation, with three contributions:
\begin{itemize}[leftmargin=*, itemsep=-0.5em, topsep=0.0em]
    \item We propose \mname that orchestrates multimodal retrieval and checklist-guided synthesis, achieving coherent multimodal generation and visual evidence grounding.

    \item We develop a rigorous trajectory curation pipeline that constructs 8K high-quality expert traces, enabling post-training of open-sourced LLMs to achieve precise multimodal selection and coherent multimodal synthesis.

    \item We design \dname for comprehensive and reproducible multimodal long-form evaluation, revealing the pressing need for advanced multimodal agentic framework and post-training.
\end{itemize}

\section{Method: Deep-Reporter}
\label{sec:method}

\subsection{The Deep-Reporter Framework}
\label{subsec:framework}

\mname (Figure~\ref{fig:deep_reporter}a) transforms a user query into a comprehensive multimodal report $\mathcal{R}^*$ via three specialized agents: a \textit{Planner}, a collaborative \textit{Searcher-Filter}, and a \textit{Reporter}.

\textit{\textbf{Sectional Planning with Dual-granularity Checklist.}}
Following prior work~\cite{shao2024assisting,prasad2024adapt}, we use hierarchical planning to maintain global coherence.
The Planner Agent $\pi_{plan}$ decomposes the user input $\mathcal{U}$ into a structured global blueprint: $\pi_{plan}(\mathcal{U}) = \{S_1, \ldots, S_N\}$, where each section $S_k = (D_k, \mathcal{C}_k)$ is formalized as a semantic checklist to enforce content rigor.
Specifically, $D_k$ denotes the coarse-grained content description, defining the scope and core topics of current section.
Meanwhile, $\mathcal{C}_k = \{c_{k,1}, \dots, c_{k,m}\}$ provides fine-grained semantic anchors, specifying facts, and arguments~\cite{bao-etal-2022-generative} to be addressed in the current section.
This dual-granularity checklist ensures both coherence and precision.
Furthermore, it allows optional human-in-the-loop refinement for better user intent alignment.

\textit{\textbf{Agentic Multimodal Search and Filtering.}}
Grounding long-form reports in verifiable evidence~\cite{yao2022react,jin2025search} requires retrieving both textual passages for factual content and visual elements (charts, diagrams, infographics). To achieve this, we implement a targeted agentic multimodal search mechanism that orchestrates dual-stream retrieval and relevance-aware filtering.
For each target section $S_k$, the searcher $\pi_{search}$ performs Decomposed Information Retrieval~\cite{trivedi2023interleaving}, formulating two complementary query streams conditioned on the sectional checklist $(D_k, \mathcal{C}_k)$ and historical context: (i) \textit{Narrative Queries} $\mathcal{Q}_{txt}$ to retrieve factual passages and statistics, and (ii) \textit{Visual Queries} $\mathcal{Q}_{img}$ to retrieve charts, diagrams, and infographics. These queries are executed against multimodal search tools, yielding a raw candidate pool $\mathcal{D}_{raw}$.

To ensure high-quality evidence, the Filter $\pi_{filter}$ applies Relevance-Aware Filtering~\cite{asai2024self}, selectively retaining passages and images that satisfy sectional constraints:
\begin{equation}
    \mathcal{E}_k = \{d \in \mathcal{D}_{raw} \mid \mathbb{I}_{ver}(d, D_k, \mathcal{C}_k) = 1\}
\end{equation}
where $\mathbb{I}_{ver}$ denotes a multimodal verification indicator. For textual evidence, we use LLMs to assess semantic entailment \textit{w.r.t.} $(D_k, \mathcal{C}_k)$. For visual evidence, a VLM evaluates the image's \textit{informativeness}, filtering out non-informative decorative images while prioritizing information-dense figures (e.g., statistical charts, technical diagrams).
This yields a concentrated evidence set $\mathcal{E}_k$.

\textit{\textbf{Multimodal Incremental Synthesis with Recurrent Context Management}.}
Generating coherent long-form multimodal reports introduces two critical context challenges: (i) maintaining narrative continuity across previously generated sections, and (ii) grounding the current section in concentrated multimodal evidence (text passages + images). Naively concatenating full history and all retrieved evidence into a single prompt leads to context overflow and degraded comprehension, particularly problematic given that images can consume thousands of tokens each.
To address this, the Reporter $\pi_{report}$ employs a Recurrent Context Update strategy~\cite{packer2023memgpt}, compressing historical context into $\mathcal{M}_{k-1} = (m_{global}, m_{local})$, where $m_{global}$ maintains a recursive summary of the narrative arc (preserving global coherence), while $m_{local}$ retains the verbatim tail of the preceding section (ensuring smooth transitions). Such a memory mechanism balances long-range dependencies with local fluency. For the $k$-th section, synthesis is formulated as a state-conditioned generation process:
\begin{equation}
    \mathcal{R}_k \sim P_\theta(\cdot \mid \underbrace{S_k}_{\text{Plan}}, \underbrace{\mathcal{E}_k}_{\text{Evidence}}, \underbrace{\mathcal{M}_{k-1}}_{\text{Memory}}, \underbrace{\rho_k}_{\text{Position}})
\end{equation}
where $\rho_k$ denotes the section's position within the global structure. After generation, the memory undergoes a update: $\mathcal{M}_k \leftarrow \Phi(\mathcal{M}_{k-1}, \mathcal{R}_k)$, propagating contextual understanding forward.

\textit{\textbf{Interleaved Multimodal Generation}.}
A key capability of \mname is seamlessly integrating visual evidence into narrative flow. To enable text-based LLMs to reason about image placement, we adopt caption-based transcription~\cite{dong2025benchmarking}, converting images in $\mathcal{E}_k$ into detailed textual descriptions. During generation,
the model learns to insert citations (e.g., \texttt{![](cite:img1)}) at contextually appropriate positions. 
% Ideally, such insertion is triggered exactly when the narrative requires visual substantiation.

\subsection{Agentic Traces Construction}
\label{subsec:training}

Current open-source LLMs remain text-centric, lacking the multimodal agentic reasoning capabilities required for our task. To bridge this gap, we construct a high-quality training corpus (Figure~\ref{fig:deep_reporter}b) using \mname (§\ref{subsec:framework}), combined with expert-in-the-loop curation and validation.

\textit{\textbf{Stage 1: Expert-in-the-Loop Planning.}}
We begin by curating a diverse seed curriculum $\mathbb{U}_{seed}$ spanning 9 domains. For each query, domain experts refine the sectional outlines and checklists $(D_k, \mathcal{C}_k)$ through iterative discussion. This expert involvement ensures that: (i) the research structure exhibits logical complexity and appropriate granularity, avoiding trivial decompositions; (ii) the checklists $\mathcal{C}_k$ contain concrete semantic anchors that require multimodal evidence; and (iii) the plans maintain diversity in both topical coverage and structural patterns.
This process results in 1K queries along with outlines and checklists.

\textit{\textbf{Stage 2: Agentic Trace Distillation.}}
Using expert-refined plans as input, we execute \mname with a frontier proprietary model to record the full multi-agent interaction traces, to effectively enhance grounded generation capabilities~\cite{bao-etal-2023-synthetic}.
We specifically focus on distilling two competencies:
\textbf{1. Agentic Multimodal Search and Filtering.} We capture the model's strategy for: (i) decomposing checklist requirements into precise and executable query streams; (ii) iteratively refining queries based on retrieved evidence quality; and (iii) distinguishing information-dense visuals from decorative or low-value images.
\textbf{2. Coherent Multimodal Synthesis}. We record the decision logic for visual citation placement within narrative flow. Specifically, the traces demonstrate: (i) \textit{when} to introduce visual evidence; (ii) \textit{where} to position citations at contextually appropriate positions to maximize image-text consistency.
Building on 1K queries, we obtain 17K agentic traces in stage 2.

\begin{table*}[t]
\centering
\renewcommand{\arraystretch}{0.8}
\setlength{\tabcolsep}{3pt}
\resizebox{1.0\linewidth}{!}{%
\begin{tabular}{l|ccc|cc|cccc|ccc}
    \toprule
    \multirow{2}{*}{\textbf{Benchmark}}
    & \multicolumn{3}{c|}{\textbf{Retrieval Environment}}
    & \multicolumn{2}{c|}{\textbf{Evi. Annotation}}
    & \multicolumn{4}{c|}{\textbf{Report Annotation}}
    & \multicolumn{3}{c}{\textbf{Task Info}} \\

    \cmidrule(lr){2-4} \cmidrule(lr){5-6} \cmidrule(lr){7-10} \cmidrule(lr){11-13}

    & \textbf{Source} & \textbf{Scale\textsuperscript{\dag}} & \textbf{Modal}
    & \textbf{Text Chk} & \textbf{Image}
    & \textbf{Outl.} & \textbf{Checkl.} & \textbf{Modal} & \textbf{\#Word}
    & \textbf{Domain} & \textbf{\#Task} & \textbf{Expert} \\
    \midrule

    \textsc{HelloBench}~\cite{shao2024assisting} & - & - & - & \xmark & \xmark & \xmark & \xmark & Text & 739 & 5 & 647 & \cxmark \\
    \textsc{LongEval}~\cite{wu2025longeval} & - & - & - & \xmark & \xmark & \cmark & \xmark & Text & 3.5k & 3 & 166 & \cmark \\
    \textsc{WritingBench}~\cite{wu2025writingbench} & - & - & - & \xmark & \xmark & \xmark & \cmark & Text & 2.7k & 6 & 1k & \xmark \\
    \textsc{LongGenBench}~\cite{wu2024longgenbench} & - & - & - & \xmark & \xmark & \xmark & \cmark & Text & 15k & 4 & 6.4k & \xmark \\
    \textsc{InterleavedBench}~\cite{liu2024holistic} & - & - & - & \xmark & \xmark & \xmark & \xmark & MM & 19 & 10 & 815 & \cxmark \\
    \textsc{LongLaMP}~\cite{kumar2024longlamp} & Static & 100 & Text & \cmark & \xmark & \xmark & \xmark & Text & 200 & 4 & 19.3k & \xmark \\

    \textsc{M-LongDoc}~\cite{chia2025m} & Static & 180 & MM & \cmark & \cmark & \xmark & \xmark & Text & 180 & 3 & 851 & \cxmark \\
    \textsc{MMDocIR}~\cite{dong2025mmdocir} & Static & 313 & MM & \cmark & \xmark & \xmark & \xmark & Text & 21 & 10 & 1.6k & \cxmark \\
    \textsc{DocBench}~\cite{zou2025docbench} & Static & 229 & Text & \xmark & \xmark & \xmark & \xmark & Text & 15 & 5 & 1.1k & \cxmark \\
    \textsc{MMDocRAG}~\cite{dong2025benchmarking} & Static & 222 & MM & \cmark & \cmark & \xmark & \xmark & Text & 24 & 10 & 4k & \cxmark \\
    \textsc{M4DocBench}~\cite{dong2025doc} & Sandbox & 304 & MM & \cmark & \cmark & \xmark & \xmark & MM & 60 & 4 & 158 & \cmark \\

    \textsc{DeepResearch}~\cite{du2025deepresearch} & Web & $\infty$ & Text & \cmark & \xmark & \xmark & \cmark & Text & 3.3k & 22 & 100 & \cmark \\

    \midrule

    \textbf{\dname (Ours)} & \textbf{Sandbox} & \textbf{61.4k} & \textbf{MM} & \textbf{\cmark} & \textbf{\cmark} & \textbf{\cmark} & \textbf{\cmark} & \textbf{MM} & \textbf{5.5k} & 9 & 247 & \cmark \\
    \bottomrule
\end{tabular}}
\vspace{-0.8em}
\caption{\textbf{\dname versus existing benchmarks.}
\textit{Evi.} denotes candidate sets for text/image retrieval.
\textit{Modality} describe the expected input/output format (MM=Multimodal).
}
\vspace{-0.8em}
\label{tab:bench_comparison}
\end{table*}

\textit{\textbf{Stage 3: Rigorous Quality Assurance}}.
Raw traces undergo strict filtering at scale using an advanced proprietary model.
\textbf{1. Retrieval Quality Control.} The verification model assesses search queries via: (i) \textit{Query Diversity}: whether the queries explore multiple facets of the checklist requirements rather than redundantly rephrasing similar searches; and (ii) \textit{Visual Alignment}: whether visual queries appropriately target the visual modalities needed for the section.
\textbf{2. Generation Quality Control.} The model aggressively flags traces exhibiting visual hallucinations~\cite{huang2024visual, bai2024hallucination} or poor multimodal integration. Specifically, it identifies: (i) \textit{Visual Selection Errors}: citing images that are factually incorrect or irrelevant; (ii) \textit{Positioning Defects}: inserting visual citations at inappropriate locations that disrupt narrative flow; (iii) \textit{Image-Text Inconsistency}: cases where textual descriptions contradict or misrepresent the cited visual content; and (iv) \textit{Image Reuse Issues}: redundantly citing the same image across sections without justification.
\textbf{3. Expert Validation.} Two domain experts independently annotated a random sample of 500 traces. The proprietary model achieved 92.4\% accuracy against expert consensus, with an inter-annotator agreement of 89.5\%. 
% These high metrics confirm the effectiveness of our quality control mechanism. 
Trajectories flagged by the automated system are discarded, distilling the raw data into 8K high-quality agentic traces.

\section{Benchmark: \dname}
\label{sec:benchmark}

As reflected in Table~\ref{tab:bench_comparison}, current benchmarks exhibit critical limitations: short multimodal responses, unimodal textual bias, and small-scale or uncontrolled retrieval environment.
To address these gaps, we introduce \dname, comprising 247 tasks across 9 domains, supported by a stable multimodal sandbox containing 95K images and 108M text chunks. As mentioned in Section~\ref{sec:introduction}, \dname enables unified multimodal assessment, fair and transparent comparison, and accessibility. Specifically, we develop a rigorous expert-in-the-loop pipeline to ensure task authenticity, sandbox realism, and annotation reliability.

\textit{\textbf{Task Collection and Curation.}} We curated a dataset of 247 high-quality tasks through expert sourcing, hybrid filtering, and structured extraction. \textbf{1. Source Selection:} Domain experts identified 100+ authoritative repositories across 9 domains, allowing us to crawl over 4,000 candidate reports. \textbf{2. Filtering:} A hybrid approach combining rule-based heuristics (e.g., token count >3k, visual richness) and LLM scoring narrowed candidates to $\sim$1,000. A subsequent manual review removed time-sensitive content to ensure high quality and domain diversity, yielding 247 final reports. \textbf{3. Processing:} We converted these reports to clean markdown using Crawl4AI~\cite{crawl4ai2024} (web) and MinerU~\cite{niu2025mineru2} (PDFs), followed by manual cleaning. Finally, we used LLM-assisted extraction with human verification to transform them into structured tasks (queries, outlines, checklists). These reports serve as task \textit{prototypes},
reflecting the open-ended and high-complexity nature of real-world multimodal research problems.

\textit{\textbf{Multimodal Sandbox Construction.}} We constructed a robust, large-scale multimodal sandbox through evidence aggregation, data ingestion, and multimodal indexing. \textbf{1. Evidence Aggregation:} We extracted citation URLs from source reports and expanded the corpus via the Google Custom Search API. For each section, we utilized an LLM to generate search queries based on task metadata, section descriptions, and checklists, retrieving the top-30 URLs per section to supplement original citations. \textbf{2. Data Ingestion:} Following URL deduplication, we conducted batch crawling and applied the previously described cleaning pipeline (Crawl4AI for HTML, MinerU for PDFs) to produce clean markdown and locally stored images. \textbf{3. Multimodal Indexing:} We performed hierarchical text chunking on the markdown content and embedded both text chunks and images using Jina Embeddings-v4~\cite{gunther2025jina}.

\textit{\textbf{Silver-Standard Annotation.}} We constructed a dual-layer annotation set evaluating both the research process and generation outcome. \textbf{1. Process-Level Evidence:} We employed a retrieval-and-verification~\cite{jin2025search} approach to build evidence pools. For each section, we utilized LLM-formulated queries to fetch candidates (Top-40 images/Top-100 text chunks per section) from the sandbox, followed by model-based relevance scoring and strict expert adjudication to eliminate false positives. This yielded verified pools averaging 168 text chunks and 103 images per task. \textbf{2. Outcome-Level Reference:} Addressing the open-ended nature of long-form generation~\cite{celikyilmaz2020evaluation,becker2024text}, we constructed Expert-Refined Silver Reports ($\mathcal{R}_{ref}$). We first utilized GPT-4.1 to synthesize drafts strictly conditioned on the verified outlines and original reports. Subsequently, domain experts performed post-editing to rectify logical inconsistencies and ensure optimal instruction adherence, establishing a rigorous baseline for comparative evaluation. See \S\ref{app:benchmark_supp} for complete benchmark details.

\begin{table*}[t]
\centering
\renewcommand{\arraystretch}{0.95}
\resizebox{0.99\linewidth}{!}{%
\begin{tabular}{l|l|ccc|ccc|ccc}
    \toprule
    \multirow{2}{*}{Method} & \multirow{2}{*}{Backbone} & \multicolumn{3}{c|}{Search} & \multicolumn{3}{c|}{Filter} & \multicolumn{3}{c}{Selection} \\
    \cmidrule(lr){3-5} \cmidrule(lr){6-8} \cmidrule(lr){9-11}
    & & Text & Image & Overall & Text & Image & Overall & Text & Image & Overall \\
    \midrule
    Na\"ive RAG & Qwen3-8B & \underline{34.9} (54.6) & 32.7 (27.3) & 33.8 (81.9) & - & - & - & 46.3 (18.7) & 7.8 (0.9) & 27.0 (19.6) \\
    Na\"ive RAG & Qwen3-32B & \textcolor{purple}{\textbf{35.0}} (54.7) & 32.8 (27.4) & 33.9 (82.0) & - & - & - & 45.7 (21.0) & 10.4 (1.0) & 28.1 (22.0) \\
    Storm-MM & Qwen3-32B & 29.5 (105.2) & 38.4 (53.8) & 34.0 (159.0) & - & - & - & 48.3 (26.3) & 12.7 (1.4) & 30.5 (27.7) \\
    \midrule
    \mname & Qwen3-8B & 29.3 (103.8) & 37.9 (52.4) & 33.6 (156.3) & 45.5 (50.4) & 50.2 (30.5) & 47.9 (81.0) & \textcolor{purple}{\textbf{49.6}} (24.5) & 8.3 (0.7) & 28.9 (25.2) \\
    \quad\textcolor{gray}{\textit{w/o Filter}} & Qwen3-8B & 29.5 (104.0) & 37.2 (52.5) & 33.4 (156.5) & - & - & - & 43.6 (31.9) & 3.8 (0.3) & 23.7 (32.2) \\
    \mname & Qwen3-8B$^\spadesuit$ & 30.9 (103.8) & \textcolor{purple}{\textbf{39.8}} (52.3) & \underline{35.3} (156.1) & \underline{46.5} (51.5) & 51.5 (31.0) & \underline{49.0} (82.5) & 46.5 (36.9) & \underline{45.0} (9.8) & \underline{45.7} (46.7) \\
    \quad\textcolor{gray}{\textit{w/o Filter}} & Qwen3-8B$^\spadesuit$ & 30.5 (104.2) & 39.6 (52.2) & 35.1 (156.4) & - & - & - & 39.2 (56.9) & 33.6 (6.3) & 36.4 (63.3) \\
    \mname & Qwen3-8B$^\clubsuit$ & 30.3 (105.6) & 38.6 (53.3) & 34.5 (158.9) & 46.2 (52.3) & 50.1 (31.3) & 48.2 (83.6) & 48.5 (22.5) & 9.8 (0.8) & 29.1 (23.4) \\
    \quad\textcolor{gray}{\textit{w/o Filter}} & Qwen3-8B$^\clubsuit$ & 31.7 (103.4) & 39.4 (52.7) & \textcolor{purple}{\textbf{35.6}} (156.1) & - & - & - & 44.2 (34.5) & 6.9 (0.5) & 25.6 (35.1) \\
    \mname & Qwen3-32B & 29.2 (106.0) & 38.1 (54.4) & 33.6 (160.4) & 45.4 (51.5) & 50.7 (30.4) & 48.0 (81.9) & 48.7 (27.6) & 19.5 (2.0) & 34.1 (29.6) \\
    \quad\textcolor{gray}{\textit{w/o Filter}} & Qwen3-32B & 29.2 (105.8) & 38.9 (54.4) & 34.1 (160.2) & - & - & - & 43.3 (32.1) & 12.1 (1.0) & 27.7 (33.1) \\
    \mname & Qwen3-32B$^\spadesuit$ & 30.9 (104.5) & 39.2 (52.9) & 35.1 (157.5) & \textcolor{purple}{\textbf{46.9}} (52.1) & \underline{51.6} (31.0) & \textcolor{purple}{\textbf{49.3}} (83.2) & 48.6 (40.3) & 41.4 (7.6) & 45.0 (47.8) \\
    \quad\textcolor{gray}{\textit{w/o Filter}} & Qwen3-32B$^\spadesuit$ & 30.7 (104.3) & 39.4 (52.5) & 35.0 (156.8) & - & - & - & 40.2 (55.2) & 33.4 (5.3) & 36.8 (60.5) \\
    \quad\textcolor{gray}{\textit{w/o Recur.}} & Qwen3-32B$^\spadesuit$ & 30.4 (103.2) & 38.7 (52.1) & 34.5 (155.3) & 46.2 (51.4) & 51.0 (30.6) & 48.7 (82.0) & 47.3 (39.1) & 39.6 (7.2) & 43.5 (46.3) \\
    \quad\textcolor{gray}{\textit{w/o Both}} & Qwen3-32B$^\spadesuit$ & 30.3 (103.0) & 38.6 (52.0) & 34.3 (155.0) & - & - & - & 38.8 (52.7) & 30.1 (5.0) & 34.5 (57.7) \\
    \mname & Qwen3-32B$^\clubsuit$ & 29.7 (105.5) & 39.4 (54.2) & 34.6 (159.7) & 45.6 (51.6) & 51.6 (31.3) & 48.6 (82.8) & \underline{49.0} (25.9) & 22.1 (2.5) & 35.5 (28.4) \\
    \mname & Llama3.3-70B & 24.6 (107.7) & 33.3 (53.4) & 29.0 (161.1) & 40.2 (52.0) & 46.2 (27.8) & 43.2 (79.8) & 42.4 (38.0) & 14.3 (2.1) & 28.3 (40.1) \\
    \mname & Llama3.3-70B$^\spadesuit$ & 30.8 (104.6) & \underline{39.6} (52.5) & 35.2 (157.1) & 45.9 (54.3) & \textcolor{purple}{\textbf{52.0}} (30.3) & 48.9 (84.6) & 47.4 (41.5) & 41.2 (6.2) & 44.3 (47.7) \\
    \mname & Llama3.3-70B$^\clubsuit$ & 24.8 (107.5) & 33.0 (53.5) & 28.9 (161.0) & 40.7 (52.1) & 45.9 (29.0) & 43.3 (81.0) & 42.3 (37.4) & 14.7 (1.9) & 28.5 (39.3) \\
    \mname & GPT-4.1 & 30.0 (105.3) & 39.5 (52.8) & 34.8 (158.1) & 45.8 (53.1) & 51.5 (31.2) & 48.6 (84.3) & 47.8 (42.8) & \textcolor{purple}{\textbf{45.4}} (7.3) & \textcolor{purple}{\textbf{46.6}} (50.0) \\
    \bottomrule
\end{tabular}}
\vspace{-0.5em}
\caption{\textbf{Evidence quality on \dname benchmark.} Each cell shows precision (\%) and quantity of evidence that are either retrieved/retained/cited.
The best score is in \textcolor{purple}{\textbf{purple boldface}} and second best is \underline{underlined}. $^\spadesuit$ and $^\clubsuit$ denote SFT and DPO respectively. \textcolor{gray}{\textit{w/o Filter}} removes the filter module; \textcolor{gray}{\textit{w/o Recur.}} removes recurrent context management; \textcolor{gray}{\textit{w/o Both}} removes both.}
\vspace{-1em}
\label{tab:deepreporter_retrieval}
\end{table*}

\section{Experiments}
\label{sec:experiments}

\subsection{Evaluation Metrics}
\label{subsec:eval_metrics}

\begin{table*}[t]
\centering
\renewcommand{\arraystretch}{0.85}
\resizebox{0.99\linewidth}{!}{%
\begin{tabular}{l|l|ccc|ccccc|ccccc|c|c}
    \toprule
    \multirow{2}{*}{Method} & \multirow{2}{*}{Backbone}
    & \multicolumn{3}{c|}{Section Anchor}
    & \multicolumn{5}{c|}{Section Content}
    & \multicolumn{5}{c|}{Full Report}
    & \multirow{2}{*}{Overall} & Report \\
    & & Desc & Check & Avg & Rich & Coh & Place & Clar & Avg & Coh & Flu & Rep & Term & Avg & & Length \\
    \midrule
    Na\"ive RAG & Qwen3-8B & 0.0 & 2.0 & 1.0 & 6.1 & 7.9 & 7.1 & 8.5 & 7.4 & 0.0 & 0.0 & 6.9 & 23.1 & 7.5 & 5.3 & 1.9k \\
    Na\"ive RAG & Qwen3-32B & 0.0 & 1.2 & 0.6 & 6.9 & 5.7 & 6.3 & 6.7 & 6.4 & 0.0 & 0.0 & 10.1 & 24.3 & 8.6 & 5.2 & 2.0k \\
    Storm-MM & Qwen3-32B & 22.3 & 25.3 & 23.8 & 7.7 & 5.7 & 7.3 & 6.5 & 6.8 & 10.1 & 20.7 & 19.4 & 14.6 & 16.2 & 15.6 & 3.2k \\
    \midrule
    \mname & Qwen3-8B & 14.2 & 19.4 & 16.8 & 8.1 & 3.6 & 6.9 & 3.6 & 5.6 & 10.5 & 23.5 & 21.9 & 18.6 & 18.6 & 13.7 & 3.9k \\
    \quad\textcolor{gray}{\textit{w/o Filter}} & Qwen3-8B & 13.0 & 14.6 & 13.8 & 8.1 & 2.0 & 6.1 & 3.2 & 4.9 & 8.1 & 21.9 & 17.8 & 20.7 & 17.1 & 11.9 & 3.8k \\
    \mname & Qwen3-8B$^\spadesuit$ & 23.9 & 26.3 & 25.1 & \underline{44.1} & \underline{36.4} & \textcolor{purple}{\textbf{40.1}} & \textcolor{purple}{\textbf{43.7}} & \underline{41.1} & 17.8 & 23.9 & 22.7 & 19.0 & 20.9 & 29.0 & 5.1k \\
    \quad\textcolor{gray}{\textit{w/o Filter}} & Qwen3-8B$^\spadesuit$ & 24.7 & 27.9 & 26.3 & 30.0 & 21.5 & 27.1 & 28.3 & 26.7 & 15.0 & 25.9 & 23.9 & 21.1 & 21.5 & 24.8 & 5.2k \\
    \mname & Qwen3-8B$^\clubsuit$ & 16.6 & 16.6 & 16.6 & 9.3 & 5.3 & 7.3 & 6.5 & 7.1 & 12.6 & 25.5 & 21.1 & 19.8 & 19.7 & 14.5 & 4.2k \\
    \quad\textcolor{gray}{\textit{w/o Filter}} & Qwen3-8B$^\clubsuit$ & 17.4 & 23.5 & 20.4 & 8.1 & 3.6 & 6.5 & 2.4 & 5.2 & 10.9 & 22.7 & 16.6 & 21.9 & 18.0 & 14.5 & 4.1k \\
    \mname & Qwen3-32B & \underline{38.1} & \underline{38.9} & \underline{38.5} & 12.2 & 10.5 & 15.4 & 10.5 & 12.2 & 24.3 & \underline{35.2} & \underline{34.8} & \underline{29.2} & 30.9 & 27.2 & 3.9k \\
    \quad\textcolor{gray}{\textit{w/o Filter}} & Qwen3-32B & 31.6 & 36.8 & 34.2 & 9.7 & 4.9 & 7.7 & 5.7 & 7.0 & 19.8 & 30.8 & 34.0 & \textcolor{purple}{\textbf{30.4}} & 28.7 & 23.3 & 4.0k \\
    \mname & Qwen3-32B$^\spadesuit$ & \textcolor{purple}{\textbf{39.7}} & \textcolor{purple}{\textbf{39.7}} & \textcolor{purple}{\textbf{39.7}} & \textcolor{purple}{\textbf{47.8}} & \textcolor{purple}{\textbf{37.2}} & \underline{38.5} & \underline{43.3} & \textcolor{purple}{\textbf{41.7}} & \underline{27.1} & \textcolor{purple}{\textbf{36.4}} & \textcolor{purple}{\textbf{36.8}} & 28.7 & \textcolor{purple}{\textbf{32.3}} & \textcolor{purple}{\textbf{37.9}} & 4.7k \\
    \quad\textcolor{gray}{\textit{w/o Filter}} & Qwen3-32B$^\spadesuit$ & 34.8 & 38.1 & 36.4 & 30.4 & 24.3 & 28.7 & 27.1 & 27.6 & 23.5 & 25.1 & 23.9 & 25.9 & 24.6 & 29.6 & 4.9k \\
    \quad\textcolor{gray}{\textit{w/o Recur.}} & Qwen3-32B$^\spadesuit$ & 36.4 & 38.8 & 37.6 & 39.2 & 30.4 & 32.8 & 34.4 & 34.2 & 11.3 & 24.3 & 25.0 & 18.6 & 19.8 & 30.5 & 4.5k \\
    \quad\textcolor{gray}{\textit{w/o Both}} & Qwen3-32B$^\spadesuit$ & 36.0 & 37.8 & 36.9 & 27.5 & 20.2 & 25.2 & 25.1 & 24.5 & 9.7 & 21.5 & 22.6 & 15.8 & 17.4 & 26.3 & 4.3k \\
    \mname & Qwen3-32B$^\clubsuit$ & 19.4 & 20.2 & 19.8 & 15.8 & 8.5 & 12.2 & 10.1 & 11.6 & 14.2 & 28.3 & 27.9 & 23.9 & 23.6 & 18.4 & 3.9k \\
    \mname & Llama3.3-70B & 4.1 & 6.5 & 5.3 & 11.3 & 6.1 & 9.3 & 6.5 & 8.3 & 4.5 & 8.1 & 10.5 & 5.7 & 7.2 & 6.9 & 3.4k \\
    \mname & Llama3.3-70B$^\spadesuit$ & 35.6 & 37.7 & 36.6 & 36.4 & \underline{36.4} & 37.7 & 36.4 & 36.7 & \textcolor{purple}{\textbf{33.2}} & 32.8 & 30.4 & 27.5 & \underline{31.0} & \underline{34.8} & 4.5k \\
    \mname & Llama3.3-70B$^\clubsuit$ & 4.9 & 7.3 & 6.1 & 11.7 & 6.5 & 10.5 & 8.9 & 9.4 & 4.9 & 13.8 & 9.3 & 7.3 & 8.8 & 8.1 & 3.5k \\
    \bottomrule
\end{tabular}}
\vspace{-0.5em}
\caption{
\textbf{Main results of generation quality.}
\textbf{Section Anchor}: Desc=Description, Check=Checklist. \textbf{Section Content}: Rich=Richness, Coh=Image-text coherence, Place=Placement, Clar=Clarity. \textbf{Full Report}: Coh=Coherence, Flu=Fluency, Rep=Repetition, Term=Termination.
$^\spadesuit$ and $^\clubsuit$ denote SFT and DPO respectively.
\textcolor{gray}{\textit{w/o Recur.}} removes recurrent context management; \textcolor{gray}{\textit{w/o Both}} removes both filtering and recurrent context.
}
\vspace{-1em}
\label{tab:deepreporter_generation}
\end{table*}

\textit{\textbf{\quad Multimodal Evidence.}}
We evaluate the process-level evidence precision of the agentic pipeline: (i) raw candidates retrieved by searcher (\textbf{Search}), (ii) concentrated candidates filtered by filter (\textbf{Filter}), and (iii) candidates selected as part of narrative flow by reporter (\textbf{Selection}).
Specifically, we report $\text{Precision@Silver}$ measuring the proportion of candidates that matches silver-standard annotations: $|\mathcal{R}^{(t)} \cap \mathcal{S}_{\text{silver}}| / |\mathcal{R}^{(t)}|$.

\textit{\textbf{Multimodal Generation.}}
We assess the long-form multimodal generation quality across three hierarchical levels: (i) structural adherence to planning constraints (\textbf{Section Anchor}), i.e., description and checklist, (ii) section-level multimodal grounding quality (\textbf{Section Content}), which measures the richness and clarity of generated section, and image-text coherence and citation format, and (iii) holistic evaluation on entire report (\textbf{Full Report}), favoring coherent and fluent writing while penalizing repetition and early termination.
Specifically, we report the \textit{Relative Quality Score}~\citep{dubois2024length,lin2024wildbench},
using LLM\&VLM to score the generated content against expert-refined silver reports $\mathcal{R}_{\text{ref}}$ and then normalize the scores (refer to \S\ref{sub:win_rate_calc} for more details).
The overall performance is computed as $(S_{\text{anchor}} + S_{\text{content}} + S_{\text{full}})/3$.

\subsection{Experimental Setup}
\label{subsec:exp_setting}

\textit{\textbf{\quad Retrieval Settings.}}
All experiments utilize the same search tool, the multimodal sandbox (\S\ref{sec:benchmark}) to ensure reproducibility.
For each search query, the tool performs semantic match and returns top-20 text chunks and top-10 images.
These budgets are chosen to balance evidence coverage against context overhead: a top-20/10 budget provides sufficient candidate diversity for the subsequent filtering stage while keeping the raw evidence volume manageable (on average $\sim$156 candidates per task, reduced to $\sim$81 after filtering; see Table~\ref{tab:deepreporter_retrieval}).
End-to-end filter runtime scales roughly linearly with the retrieval budget, allowing users to trade off quality and latency by adjusting these parameters.

\textit{\textbf{\mname.}}
We implement our framework using Qwen3-8B/32B~\citep{yang2025qwen3} and Llama-3.3-70B-Instruct~\citep{dubey2024llama}:
\begin{itemize}[leftmargin=*, itemsep=-0.3em, topsep=0.0em]
    \item \textbf{Base} utilizes the agentic workflow via in-context learning without weight updates.
    \item \textbf{SFT} performs fine-tuning on 8K agentic traces (\S\ref{subsec:training}) for agentic search and report generation.
    \item \textbf{DPO} applies preference learning \citep{rafailov2023direct} using 8K agentic trace (details in \S\ref{sub:dpo_config}).
    \item \textbf{w/o Filter} is an ablation variant that removes filter to quantify the impact of noise reduction.
    \item \textbf{w/o Recur.} removes recurrent context management while keeping filter, isolating the contribution of cross-section memory.
    \item \textbf{w/o Both} removes both filter and recurrent context, quantifying their combined effect.
\end{itemize}

\textit{\textbf{Na\"ive RAG Baselines.}}
Rather than using agentic search, we compare \mname against na\"ive RAG baselines using a non-iterative retrieval strategy.
Given the sectional anchors, RAG baselines retrieve top-10 text chunks and top-5 images per section in parallel.
The retrieved candidates are subsequently fed to backbone models for answer synthesis, without any relevance check or noise filtering.

\textit{\textbf{Storm-MM Baseline.}}
To provide a stronger agentic baseline, we construct Storm-MM, a minimally adapted STORM-style~\citep{shao2024assisting} agent under the same controlled environment as \mname (same backbone LLM, same retrieval tool and sandbox, same evaluation metrics).
We retain STORM's core multi-turn research pipeline and make three necessary changes to enable multimodal output: (i) the agent issues both text and image retrieval queries during multi-turn research, (ii) retrieved text chunks and image captions are merged into a unified multimodal evidence pool, and (iii) the section-level writer is prompted to generate output with both text and image citations.

\subsection{Evidence Quality Analysis}
\label{subsec:res_retrieval}

Table~\ref{tab:deepreporter_retrieval} presents the multimodal evidence quality at three stages of the agentic pipeline.

\textit{\textbf{Raw multimodal search yields comparable evidence quality across methods.}}
Under identical retrieval settings, raw search performance remains largely consistent across both textual and visual modalities.
For example, with the Qwen3-32B backbone, na\"ive RAG achieves a search precision of 33.9 in overall, while \mname (Base) attains a comparable score of 33.6. This consistency confirms that all methods operate under a stable and fair retrieval environment, removing the impact of raw retrieval on downstream modules.

\textit{\textbf{Agentic multimodal search and filtering substantially concentrates evidence.}}
Despite similar raw retrieval quality, \mname achieves significant gain on evidence precision after the filter stage, across all backbones. With Qwen3-8B, overall precision increases from 33.6 to 47.9 after filtering,
while retaining a substantial volume of evidence (81.0), which markedly reduces the scale of multimodal context compared to raw retrieval.
These results show that the agentic \textit{Searcher–Filter} design effectively transforms noisy multimodal retrieval outputs into a concentrated evidence set optimized for downstream usage.

\textit{\textbf{Training activates reliable multimodal evidence selection.}}
The training variants consistently outperform their base counterparts for evidence selection,
especially for images. For Qwen3-8B, training increases image Selection precision from 8.3 to 45.0, while expanding the number of cited images from 0.7 to 9.8 per report (13x improvement). This suggests that training does not merely refine retrieval behavior, but actively teaches the model to identify and integrate information-dense visual evidence into the narrative flow, enabling effective multimodal grounding in long-form generation.

\subsection{Main Results: Multimodal Generation}
\label{subsec:res_generation}

Table~\ref{tab:deepreporter_generation} validates our core contributions via systematic comparisons.

\textit{\textbf{The agentic workflow enables coherent multimodal long-form generation.}}
Na\"ive RAG collapses to short and shallow outputs: with Qwen3-32B, it produces reports of only 2.0k tokens and achieves a low overall score of 5.2. In contrast, \mname (Base) generates substantially longer reports (3.9k tokens) and reaches an overall score of 27.2, corresponding to a 4.2× improvement in generation quality.
This gain reflects a consistent improvement across structural, multimodal, and holistic dimensions. \mname (Base) exhibits strong adherence to section anchors (Section Anchor Avg: 0.6 → 38.5), substantially richer and more coherent multimodal content (Section Content Avg: 6.4 → 12.2), and markedly improved full-report coherence and fluency (Full Report Avg: 8.6 → 30.9).
To further test whether the gains come from framework design rather than simply from iterative search, we compare against Storm-MM (\S\ref{subsec:exp_setting}), a stronger STORM-style iterative agentic baseline. Although Storm-MM already surpasses Na\"ive RAG (overall: 5.2 $\rightarrow$ 15.6), \mname (Base) still leads by a large margin (overall: 27.2 vs.\ 15.6; Sel.\ Overall: 34.1 vs.\ 30.5).
Together, these results show that the agentic workflow of \mname can successfully transform multimodal retrieval results into structured, visually grounded long-form reports.

\textit{\textbf{Post-training can boost multimodal generation in all dimensions.}}
Building on a strong agentic base, post-training yields further substantial improvements in section anchor and content, and overall report.
\textbf{1. For backbones that perform poorly in the base setting:} these gains are broad and pronounced: Llama3.3-70B improves from an overall score of 6.9 to 34.8 (over $5\times$), reflecting large and consistent improvements across structural adherence (6.9x), multimodal content quality (4.4x), and full-report coherence (4.3x) rather than isolated metric gains.
\textbf{2. For stronger backbones:} post-training further strengthens performance across all dimensions while preserving generation stability.
On Qwen3-32B, training increases the overall score from 27.2 to 37.9 (+39\%) and extends report length from 3.9k to 4.7k tokens, with the most pronounced improvement observed in multimodal content, where Section Content Avg rises from 12.2 to 41.7.
Together, these results show that post-training consistently enhances multimodal grounding and long-form synthesis quality by activating missing capabilities in weaker models and boosting existing strengths in stronger ones.

\subsection{Ablation Analysis}
\label{subsec:ablation}

We conduct ablations to isolate the effects of filtering, recurrent context management, model scale, and training objectives.

\textit{\textbf{Filtering stabilizes multimodal generation by reducing context noise.}}
Removing the Filter module consistently degrades generation quality, with the largest drops in section-level multimodal content.
For Qwen3-32B$^\spadesuit$, overall performance decreases from 37.9 to 29.6, while Section Content Avg drops sharply from 41.7 to 27.6.
Specifically, raw retrieval yields over 150 candidates on average, whereas filtering compresses the evidence set to around 80 items, concentrating relevance and alleviating context overload.
These results indicate that filtering prevents context pollution that disrupts image--text integration and citation decisions, rather than merely improving retrieval metrics.
Notably, filtering dominates runtime ($\sim$70\% of total latency, Appendix~\ref{sub:runtime_analysis}); post-training, in contrast, adds no inference overhead, making all quality gains from SFT essentially free.

\textit{\textbf{Recurrent context management is essential and complementary to filtering.}}
To isolate the impact of recurrent context management (\S\ref{subsec:framework}), we evaluate two additional ablations on Qwen3-32B$^\spadesuit$: \textit{w/o recurrent ctx} removes the recurrent context update while keeping filtering, and \textit{w/o both} removes both modules simultaneously.
Removing recurrent context management causes the largest drop in Full Report quality (32.3 $\rightarrow$ 19.8), indicating that the long-context LLM alone cannot maintain cross-section coherence without explicit context accumulation.
In contrast, removing the filter produces the largest drop in Section Content (41.7 $\rightarrow$ 27.6) and selection precision (Sel.\ Overall: 45.0 $\rightarrow$ 36.8 in Table~\ref{tab:deepreporter_retrieval}).
Removing both modules compounds the degradation (overall: 37.9 $\rightarrow$ 26.3; Sel.\ Overall: 45.0 $\rightarrow$ 34.5), confirming that these two components address complementary challenges: recurrent context for long-range coherence vs.\ filtering for evidence quality.

\begin{figure}[t]
    \centering
    \includegraphics[width=1.0\linewidth]{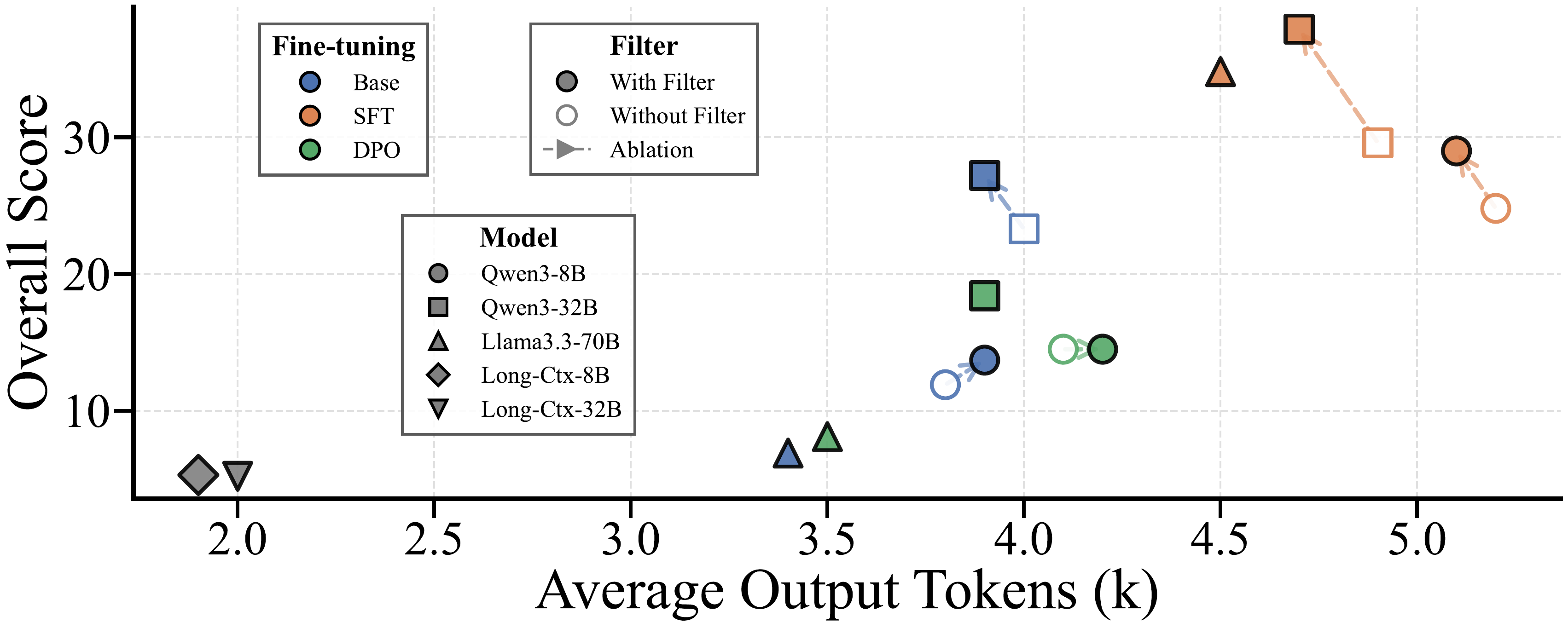}
    \vspace{-1.5em}
    \caption{Overall performance with output tokens.}
    \vspace{-2.0em}
    \label{fig:scatter_plot}
\end{figure}

\textit{\textbf{Model scale raises the ceiling, but agentic orchestration drives the qualitative shift.}}
Larger backbones yield higher performance once the agentic workflow is in place, but scale alone is insufficient. Under Na\"ive RAG, Qwen3-32B underperforms Qwen3-8B (5.2 vs 5.3), whereas with \mname, Qwen3-32B substantially outperforms Qwen3-8B both before (27.2 vs 13.7) and after training (37.9 vs 29.0). Figure~\ref{fig:scatter_plot} shows that, at comparable lengths, larger models occupy higher-quality regions. Overall, scale amplifies performance, while agentic orchestration enables the transition from short summaries to grounded long-form reports.

\textit{\textbf{Supervision granularity determines the effectiveness of training objectives.}}
SFT consistently outperforms DPO for long-form multimodal generation. For Qwen3-32B, SFT achieves 37.9 overall versus 18.4 for DPO, with a large gap in Section Content (41.7 vs 11.6); similar trends hold for Qwen3-8B. This difference is consistent with supervision granularity: image selection and placement are sparse decisions in long trajectories, for which token-level SFT provides direct learning signals, while trajectory-level preference optimization offers weaker supervision (more details in \S~\ref{app:add_results}).

\textit{\textbf{Generalization to open web.}}
To validate that findings on our sandbox transfer to an open environment, we conduct a small-scale experiment: for 100 tasks (Qwen3-32B backbone), we replace the sandbox retrieval tool with the Google Search API and re-run \mname on the augmented evidence pool.
Results show that 83.6\% of URLs returned by live Google Search are already present in our sandbox, confirming that the corpus construction effectively covers what the open web provides for these tasks.
Moreover, \mname maintains comparable generation quality on live web evidence (Gen.\ Overall: 26.2 vs.\ 27.2 on sandbox), with no significant degradation in Section Anchor (40.3 vs.\ 38.5) or Full Report quality (28.6 vs.\ 30.9).
These results demonstrate that the retrieval component is modular and readily replaceable, and that the factual grounding is preserved even in an open environment.

\section{Related Work}
\label{sec:related_work}

\textit{\textbf{\quad From RAG to Autonomous Research Agents}.}
RAG~\cite{lewis2020retrieval} and ReAct~\cite{yao2022react} underpin modern agentic systems.
Recent work extends them to autonomous, multi-step research agents, where explicit planning improves search trajectories~\cite{zheng2025deepresearcher,xue2025simpletir,hu2025flowsearch,shi2025deepdiver0,zhang2026srr0judge0}, search decision marking~\cite{zhang2026das}.
However, existing agents remain predominantly \textit{text-centric}~\cite{geng2025webwatcher,lin2025ragcapbenchbenchmarkingcapabilitiesllms}: visual artifacts are linearized or treated as weak retrieval cues, losing their structured semantics~\cite{yu2024visrag}, and rarely serve as first-class evidence in the reasoning loop~\cite{wasserman2025real, abootorabi2025ask}.
While multimodal RAG pipelines such as MMDocRAG~\cite{dong2025benchmarking} and M-LongDoc~\cite{chia2025m} introduce cross-modal retrieval, they target single-round document-level QA with short outputs, which differs fundamentally from the iterative, multi-section long-form generation addressed by our work.

\textit{\textbf{Multimodal Long-Form Generation.}}
Current systems face a fundamental trade-off between multimodality and long-range generation.
Large Multimodal Models perform well on short interleaved reasoning~\cite{lillava, du2025easy} but degrade at document scale, while long-form text generators achieve stable 10k+ outputs~\cite{bai2024longwriter, gu2025rapid, he2025precise, guo2025general} without grounded visual evidence.
Although recent agentic approaches synthesize charts or multimodal reports~\cite{li2025metal, kaur2025chartagent, yang2025multimodal}, they favor synthetic presentation over evidence fidelity, leaving authentic visual artifacts unsupported as first-class evidence.

\section{Conclusion}
\label{sec:conclusion}

We present \mname, a unified agentic framework that advances long-form generation from text-centric to truly multimodal research capabilities. Through agentic multimodal search and filtering, checklist-guided incremental synthesis, and recurrent context management, our framework enables coherent integration of textual and visual evidence in long-form generation. We construct 8K high-quality agentic traces via expert-in-the-loop curation.
Our comprehensive benchmark, \dname, establishes a rigorous and reproducible testbed with 247 research tasks and a stable multimodal sandbox.
Extensive experiments reveal that relevance-aware filtering and recurrent context management are complementary components critical for evidence quality and long-range coherence, and that training with curated agentic trajectories activates multimodal selection and integration capabilities.
\mname also generalizes to the open web.
These findings validate the effectiveness of \mname and the value of our data curation pipeline.
We hope \mname serves as a foundation for future research in grounded, multimodal content creation.

\clearpage
\section*{Limitations}

Despite strong performance, our framework has several limitations:

\paragraph{Static Multimodal Sandbox.} While our sandbox ensures fairness, transparency, and reproducibility through its controlled environment, it remains static by design. This presents challenges when users need to explore topics outside the predefined domains. However, our sandbox is inherently scalable. We encourage users to leverage our curation pipeline to expand the corpus using seed topics before performing deep research on brand-new out-of-domain subjects.

\paragraph{Offline Optimization Only.} Our current optimization is limited to supervised fine-tuning (SFT) and direct preference optimization (DPO), without exploring online reinforcement learning methods. We conducted preliminary experiments with GRPO but found limited success, which we attribute to the inherent difficulty of open-ended RL in settings where rewards are sparse and hard to determine. We encourage future work to investigate more sophisticated online RL approaches tailored for multimodal long-form generation.

\paragraph{English-Only Focus.} Our framework and benchmark currently support English only. Extending \mname to multilingual settings remains an important direction for future work, particularly given the growing demand for research capabilities across diverse languages and cultural contexts.

\section*{Ethical Considerations}

\paragraph{Intended use and usage constraints.}
The benchmark and datasets introduced in this work are intended solely for academic research, including the evaluation and analysis of multimodal agentic retrieval systems. They are not designed for real-world deployment, commercial use, or high-stakes applications. All benchmark components are constructed in accordance with the intended use and license conditions of the underlying resources. When source artifacts are restricted to research-only or non-commercial use, all derived data inherit the same constraints.

\paragraph{Bias and representation.}
Retrieved multimodal evidence may reflect biases present in underlying data sources, including imbalanced geographic, cultural, or demographic representation.

\paragraph{Privacy and sensitive content.}
Our data curation pipeline avoids private or personally identifiable information and relies exclusively on publicly available, non-personal research-oriented materials. The benchmark tasks focus on analytical and explanatory capabilities rather than personal profiling or surveillance-related use cases.

\paragraph{Use of AI assistants.}
Large language models were used as auxiliary tools during benchmark construction, including query formulation and draft synthesis for silver annotations. Most AI-generated outputs were reviewed and refined by human experts, who remained fully responsible for task design, annotation, and the final conclusions. ChatGPT was additionally used to improve writing quality and presentation.

\section*{Acknowledgements}
We thank the anonymous reviewers and the area chair for their constructive feedback, which significantly improved this work. We also thank the domain experts who contributed to task curation, outline refinement, and silver annotation verification.

\bibliography{custom}
\clearpage
\appendix
\section*{Appendix Overview}
\label{app:overview}

This appendix provides detailed supplementary materials supporting the contributions of \mname. The content is strictly organized as follows:

\begin{itemize}[leftmargin=*, label=$\bullet$, itemsep=0em, topsep=-0.1em]
    \item \textbf{Appendix \ref{app:benchmark_supp}: \dname Construction Details.}
    \begin{itemize}[leftmargin=*, itemsep=0em, topsep=-0.1em]
        \item \textbf{\S\ref{subsec:benchmark_imp}}: Sandbox Construction and Implementation Details.
        \item \textbf{\S\ref{subsec:benchmark_statistics}}: Benchmark Statistics.
        \item \textbf{\S\ref{sub:bench_resources}}: Analysis of human labor, time investment, and computational resources.
        \item \textbf{\S\ref{sub:bench_prompts}}: The specific LLM and VLM prompts used for data curation and filtering.
        \item \textbf{\S\ref{sub:annotation_flow}}: Detailed visual breakdown of the annotation workflow.
    \end{itemize}

    \item \textbf{Appendix \ref{app:exp_details}: Experimental Setup Details.}
    \begin{itemize}[leftmargin=*, itemsep=0em, topsep=-0.1em]
        \item \textbf{\S\ref{sub:model_versions}}:Specific model versions, revision hashes and licenses.
        \item \textbf{\S\ref{sub:system_config}}: Detailed system configuration and more environment settings.
        \item \textbf{\S\ref{sub:framework_prompts}}: Full prompts for Planner, Searcher, Filter, and Reporter agents.
    \end{itemize}

    \item \textbf{Appendix \ref{app:training_details}: Training Implementation.}
    \begin{itemize}[leftmargin=*, itemsep=0em, topsep=-0.1em]
        \item \textbf{\S\ref{sub:sft_params}}: SFT hyperparameters and compute resource usage.
        \item \textbf{\S\ref{sub:dpo_config}}: DPO dataset construction methodology and training parameters.
    \end{itemize}

    \item \textbf{Appendix \ref{app:eval_details}: Evaluation Details.}
    \begin{itemize}[leftmargin=*, itemsep=0em, topsep=-0.1em]
        \item \textbf{\S\ref{sub:scoring_details}}: Detailed scoring rubrics.
        \item \textbf{\S\ref{sub:win_rate_calc}}: Final Score calculation logic.
        \item \textbf{\S\ref{sub:human_agreement}}: Human-LLM alignment study and discussion.
        \item \textbf{\S\ref{sub:eval_cost}}: Breakdown of evaluation costs.
    \end{itemize}

    \item \textbf{Appendix \ref{app:add_results}: Additional Experimental Results.}
    \begin{itemize}[leftmargin=*, itemsep=0em, topsep=-0.1em]
        \item \textbf{\S\ref{sub:runtime_analysis}}: Analysis of generation latency and inference costs.
        \item \textbf{\S\ref{sub:radar_domain}}: Capability Radar Charts and fine-grained Domain Performance tables.
        \item \textbf{\S\ref{sub:sft_dpo}}: Analysis of DPO Underperformance.
    \end{itemize}

    \item \textbf{Appendix \ref{app:case_studies}: Qualitative Case Studies.}
    \begin{itemize}[leftmargin=*, itemsep=0em, topsep=-0.1em]
        \item Display of 3 generated report samples.
    \end{itemize}
\end{itemize}

\newpage
\section{\dname Construction Details}
\label{app:benchmark_supp}

\begin{figure*}[ht]
    \centering
    \includegraphics[width=0.98\linewidth]{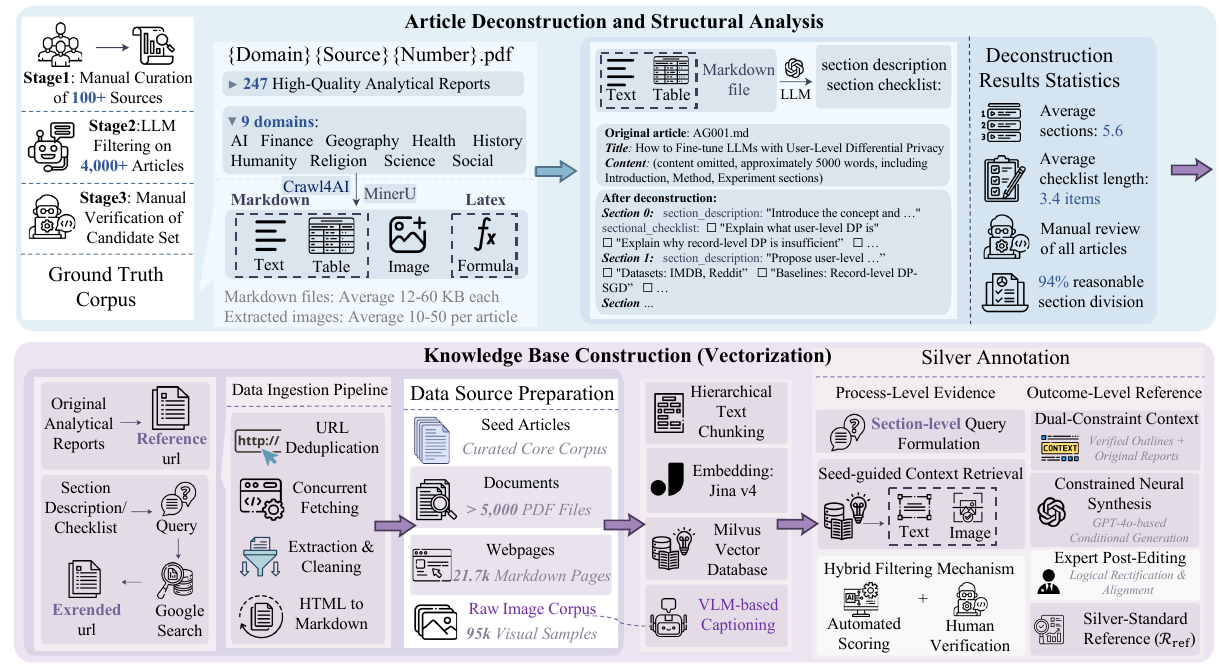}
    \caption{Demonstration of the sandbox construction.}
    \label{fig:deep_reporter_bench}
\end{figure*}

\subsection{Sandbox Construction and Implementation Details}
\label{subsec:benchmark_imp}

\textit{\quad \textbf{More Details.}}
We implement the sandbox as a multimodal retrieval and grounding pipeline (Figure~\ref{fig:deep_reporter_bench}), with a modular design similar to \cite{dong2025doc}.

We start from curated analytical reports and their reference URLs, and run a lightweight ingestion pipeline (URL deduplication, concurrent fetching, extraction/cleaning) to normalize heterogeneous sources. For web content, we convert pages into Markdown to preserve structural cues (e.g., headings, lists, tables) while keeping links to the original sources.

For PDF reports, we convert them into Markdown using MinerU v2.0.6, and apply the same processing pipeline used for web-extracted Markdown content. Textual content is segmented at subheading boundaries and each paragraph is retained to maintain semantic integrity. Short paragraphs are merged until the length approaches but not exceeds 350 words. Images smaller than 10KB are removed and each visual chunk retains its page index and layout bounding box for localization.

We embed all chunks and images using jina-embeddings-v4 \cite{jin2025search}. To improve text-to-visual access and cross-modal alignment, we caption every visual chunk with InternVL3.5-38B \cite{wang2025internvl3}. We index embeddings in Milvus for dense top-$K$ retrieval, and store metadata and structured artifact in MySQL. During retrieval, image candidates are returned with their original file paths, captions, as well as base64-encoded visual contents, allowing downstream models to directly ingest images for further cross-modal reasoning and generation.

\textit{\textbf{Sandbox Extensibility.}}
Beyond the curated sources used in this release, the \dname\ sandbox is designed to be \emph{incrementally extensible}.
Given a user-specified \emph{seed topic} (optionally with a small set of seed URLs), we provide an automated discovery script that performs topic-guided search, URL harvesting, and controlled web crawling to collect candidate evidence pages and reports for future sandbox expansion.
The script applies the same normalization and deduplication rules as our ingestion pipeline, preserves source provenance (original URLs and link graph metadata), and supports incremental indexing into our storage (MySQL) and dense retrieval backend (Milvus).
This tooling makes it straightforward to grow the sandbox over time with new domains and emerging topics while maintaining a stable, transparent, and reproducible retrieval environment.

\textit{\textbf{Personally Identifiable and Offensive Content.}}
All data sources are publicly accessible authoritative repositories intended for open information dissemination. During data collection and processing, we explicitly filter out personally identifiable information (PII), sensitive personal data, and offensive or harmful content. The curated tasks and sandbox materials focus exclusively on technical, scientific, and analytical content and do not involve private individuals or user-generated data.

\subsection{Benchmark Statistics}
\label{subsec:benchmark_statistics}

Table~\ref{tab:bench_stat} provides a comprehensive statistical breakdown of \dname. The data quantitatively validates our design choices in addressing the limitations of prior benchmarks:

\textit{\textbf{Sandbox Sufficiency: Replacing the Live Web.}}
A core contribution of \dname is decoupling long-form research from the instability and opacity of live search engines without sacrificing information density. As shown in the \textit{Sandbox Scale} columns of Table~\ref{tab:bench_stat}, we construct a retrieval environment that is informationally complete. This scale ensures that the sandbox contains sufficient "distractor" noise and diverse evidence to mimic real-world discovery, effectively serving as a stable, transparent, and accessible alternative to commercial web search APIs.

\textit{\textbf{Multimodal Complexity.}}
To bridge the gap in multimodal long-form generation, \dname presents tasks of significant depth. The \textit{Source Report Metadata} in Table~\ref{tab:bench_stat} indicates that the ground-truth blueprints average 5.5k tokens and 9.5 images. This complexity compels agents to move beyond simple factoid retrieval, requiring them to synthesize narrative structures and integrate visual evidence coherently across lengthy documents.

\textit{\textbf{Structural Granularity for Rigorous Evaluation.}}
Finally, to support our \textit{Unified Multimodal Assessment}, the dataset features high structural specificity. The \textit{Tasks} columns reveal that each report is decomposed into an average of 5.6 sections, guided by 3.4 fine-grained checklist constraints per section. This density of explicit instructions (totaling $\sim$19 constraints per task) transforms the vague "open-ended generation" problem into a measurable process, enabling precise adherence checking against the silver-standard annotations (167.9 chunks and 102.6 images per report).

\begin{table*}[ht]
\centering
\renewcommand{\arraystretch}{0.95}
\resizebox{0.99\linewidth}{!}{%
\begin{tabular}{l|ccc|cc|cccc|cc}
    \toprule
    \multirow{2}{*}{\textbf{Domain}} & \multicolumn{3}{c|}{\textbf{Source Report Metadata}} & \multicolumn{2}{c|}{\textbf{Tasks}} & \multicolumn{4}{c|}{\textbf{Sandbox Scale (Original / Expanded)}} & \multicolumn{2}{c}{\textbf{Silver Annot.}} \\
    \cmidrule(lr){2-4} \cmidrule(lr){5-6} \cmidrule(lr){7-10} \cmidrule(lr){11-12}

    & Count & Image & Token & Sec. & Check. & Web Pages & PDF Docs & Images & Chunks  & Images & Chunks \\
    \midrule

    AI & 43 & 12.2 & 7.3k & 5.7 & 3.6 & 28.5 / 59.8 & 16.8/ 27.9  & 39.7 / 289.9 & 103.0k / 179.3k & 98.7 & 170.2 \\
    Finance & 37 & 10.8 & 7.5k & 5.9 & 3.4 & 16.4 / 79.5 & 1.7/ 13.3  & 20.8 / 285.9 & 59.3k / 238.2k & 110.5 & 176.0 \\
    Geography & 12 & 8.8 & 4.1k & 5.6 & 3.0 & 8.3 / 80.1 & 2.0 / 9.5  & 45.6 / 548.3 & 29.9k / 240.0k & 99.7 & 165.3 \\
    Health & 28 & 7.0 & 4.2k & 5.8 & 3.3 & 19.2 / 77.0 &  5.5 / 6.7  & 45.3 / 287.5 & 70.0k / 230.6k & 108.5 & 172.5 \\
    History & 22 & 7.4 & 3.3k & 5.1 & 3.1 & 3.3 / 59.0 & 1.5 / 11.1  & 8.3 / 592.6 & 23.0k / 324.1k & 99.1 & 154.0 \\
    Humanity & 34 & 6.5 & 4.1k & 5.1 & 3.4 & 11.3 / 62.6 & 3.6 / 7.3  & 26.0 / 263.3 & 199.1k / 911.2k & 92.3 & 151.8 \\
    Religion & 15 & 10.5 & 5.5k & 6.1 & 3.5 & 17.4 / 74.0 & 1.8/ 5.8  & 66.4 / 342.5 & 63.0k / 221.7k & 116.3 & 183.9 \\
    Science & 41 & 10.3 & 5.6k & 5.4 & 3.2 & 17.7 / 77.0 & 2.6 / 10.9  & 58.8 / 387.9 & 66.4k / 230.7k & 96.7 & 162.4 \\
    Social & 15 & 9.6 & 5.1k & 6.4 & 3.7 & 20.0 / 78.4 & 2.4 / 9.2  & 73.3 / 263.9 & 72.4k / 234.9k & 115.8 & 190.0 \\

    \textbf{Overall} & 247 & 9.5 & 5.5k & 5.6 & 3.4 & 17.1 / 70.9 & 7.6 / 12.8  & 41.8 / 342.8 & 82.9k / 313.2k & 102.6 & 167.9 \\

    \bottomrule
\end{tabular}}
\caption{\textbf{Statistical overview of \dname.} \textbf{Source Report Metadata} reflects the complexity of the ground-truth blueprints. \textbf{Sandbox Scale} highlights the difficulty of the retrieval task by contrasting original citations with our significantly expanded search space (Expanded), demonstrating the ``needle-in-a-haystack'' nature of the benchmark.}
\label{tab:bench_stat}
\end{table*}

\subsection{Resource and Cost Analysis}
\label{sub:bench_resources}

The construction of \dname required substantial investment across human expertise, computational infrastructure, and API services. The resource breakdown is detailed as follows:

\begin{itemize}[leftmargin=*, itemsep=0em, topsep=-0.1em]
    \item \textbf{Human Labor:}
     Human participants were recruited via open and transparent channels (e.g., poster advertisements), primarily consisting of senior undergraduate students, graduate students, and Ph.D. candidates with relevant academic backgrounds. We devoted approximately 420 person-hours to benchmark construction and verification. This includes expert-driven task curation and topic selection (approximately 120 hours), manual review and refinement of section outlines and checklists (approximately 120 hours), and silver annotation with relevance verification across candidate textual and visual evidence, as well as manual review and refinement of the silver reports (approximately 180 hours). All participants were compensated at a rate of approximately \$20 per hour. Human supervision was critical for ensuring task diversity, outline correctness, and high-quality multimodal relevance judgments.
    \item \textbf{Data Processing:} Computational preprocessing was conducted in an offline and fully reproducible manner. Specifically, we processed 45K webpages using Crawl4AI (approximately 36 CPU-hours), converted 6.4K PDFs using MinerU (approximately 220 GPU-hours), generated captions for 95K images (approximately 310 GPU-hours).
    All processing steps were performed once and reused across experiments.
    \item \textbf{API Usage (GPT-4.1).} We utilized GPT-4.1 primarily for query decomposition, outline validation, and relevance scoring during silver annotation. In total, we consumed approximately 18M tokens for query and outline generation and 26M tokens for relevance assessment and verification.
\end{itemize}

The total duration of the construction pipeline, from initial crawling to final indexing, was approximately 3 months.

\subsection{Prompts for Data Curation}
\label{sub:bench_prompts}
We provide the core prompts used in the data pipeline below.

\subsection{Annotation Workflow Visualization}
\label{sub:annotation_flow}
As described below, the annotation workflow proceeds through four stages: information extraction, article deconstruction, augmented search with embedding, and retrieval-driven silver annotation.

\onecolumn
    \begin{promptbox}[title={Prompt Template: Structured Table \& Figure Decomposition}]
\begin{lstlisting}[style=promptstyle]
Given an image containing a table or figure, please provide a structured and detailed description with two levels of granularity:

Coarse-grained Description:
- Summarize the overall content and purpose of the image.
- Briefly state what type of data or information is presented (e.g., comparison, trend, distribution).
- Mention the main topic or message conveyed by the table or figure.

Fine-grained Description:
- Describe the specific details present in the image.
- For tables: List the column and row headers, units, and any notable values, patterns, or anomalies.
- For figures (e.g., plots, charts): Explain the axes, data series, legends, and any significant trends, outliers, or data points.
- Note any labels, captions, or annotations included in the image.
- Highlight specific examples or noteworthy details.

Deliver the description in a clear, organized, and reader-friendly manner, using bullet points or paragraphs as appropriate.
\end{lstlisting} 
\end{promptbox}

    \begin{promptbox}[title={Prompt Template: Semantic Article Deconstruction \& JSON Structuring}]
\begin{lstlisting}[style=promptstyle]
<system_role>
You are a meticulous and precise Technical Content Analyst. Your expertise lies in faithfully deconstructing written articles into a structured format that accurately preserves the original author's intent, structure, and flow of topics. Your output must be a clean, precise representation of the source material.
</system_role>

<user_prompt>
**Primary Goal**
Your task is to analyze the provided article in Markdown format and convert it into a structured JSON object. It is crucial that the generated outline and checklists **faithfully reflect the structure, sequence, and key points of the original article**.

**Input Article**
<article_content>
{article_markdown_content}
</article_content>

**Your Cognitive Process (Instructions)**
You MUST follow this structured thinking process:

1.  **De-noise Content**: First, mentally filter out and ignore all non-essential elements: advertisements, self-promotion, conversational filler ("Hey everyone," "Thanks for reading"), website navigation elements, and any metadata. Focus only on the substantive content that advances the main argument or narrative.

2.  **Infer Core Intent**: Based on the core content, determine the central topic to formulate the `query` and identify the key promises or questions answered to create the `overall_checklist`.

3.  **Identify Natural Content Divisions**:
   - **CRITICAL**: Completely ignore markdown heading hierarchy (#, ##, ###) as it may be corrupted or inconsistent due to web scraping issues.
   - Instead, read through the entire article and identify natural logical breaks where the discussion shifts to a substantially different aspect of the main topic.
   - Look for transition phrases, topic changes, shifts in perspective, or movement from one major concept to another.
   - **Target 3-8 sections total** - aim for meaningful, substantial sections rather than many small ones.
   - Each section should represent a complete thought or major component of the overall argument.

4.  **Section Boundaries Guidelines**:
   - A new section should only begin when there's a clear shift in focus, not just a new paragraph or minor sub-point.
   - Consider combining related content that might appear under separate headings if they serve the same logical purpose.
   - Look for natural narrative flow: Introduction -> Main Arguments/Evidence -> Analysis -> Conclusions, or Problem -> Methods -> Results -> Implications, etc.
   - Avoid creating sections that are too short (less than 2-3 substantial paragraphs) or too numerous (more than 8 sections).

5.  **Section Description and Sub-topics**: For each identified section:
   - Write a comprehensive `section_description` that captures the main thrust and purpose of that section.
   - Only include sub-topics if the section genuinely covers multiple distinct concepts that would benefit from enumeration.
   - Format sub-topics as: "Key sub-topics include: 1. Topic A details, 2. Topic B details, 3. Topic C details" - but only when this truly adds clarity.

6.  **Checklist Allocation**: Distribute the `overall_checklist` items across sections based on where those concepts are actually discussed in the article, ensuring comprehensive coverage without redundancy.

**Strict Output Specification**
Your output MUST be a single, valid JSON object and nothing else. The JSON object must follow this exact structure:

<output_format>
{{
  "query": "A concise query that accurately reflects the central topic of the article.",
  "overall_checklist": [
    "A key question the original article answers.",
    "A core concept the original article explains.",
    "A main conclusion or takeaway from the original article."
  ],
  "detailed_outline": [
    {{
      "section_description": "A detailed summary of what this section covers. **If, and only if, it is beneficial for clarity,** this field should also contain a list of key sub-topics. You can **include, create, or completely omit** sub-topics based on whether they genuinely help in structuring the section's content. For example, a simple section may only require a summary paragraph. A more complex section might look like: 'This section details the experiment results. Key sub-topics include: 1. Result A ..., 2. Result B...'.",
      "sectional_checklist": [
        "A specific point or claim made within this section of the original article.",
        "A definition provided in this section."
      ]
    }},
    {{
      "section_description": "A detailed summary for the next section, following the same rules.",
      "sectional_checklist": [
        "Another specific point covered in this part of the original article."
      ]
    }}
  ]
}}
</output_format>

**Critical Constraints**
- Generate between 3-8 sections maximum
- Focus on logical content flow rather than markdown structure
- Each section must be substantial and meaningful
- Ensure sections follow the natural progression of the author's argument
- Do not create micro-sections for minor points

</user_prompt>
\end{lstlisting} 
\end{promptbox}

    \clearpage
\twocolumn

\clearpage
\begin{center}
    \includegraphics[width=\textwidth]{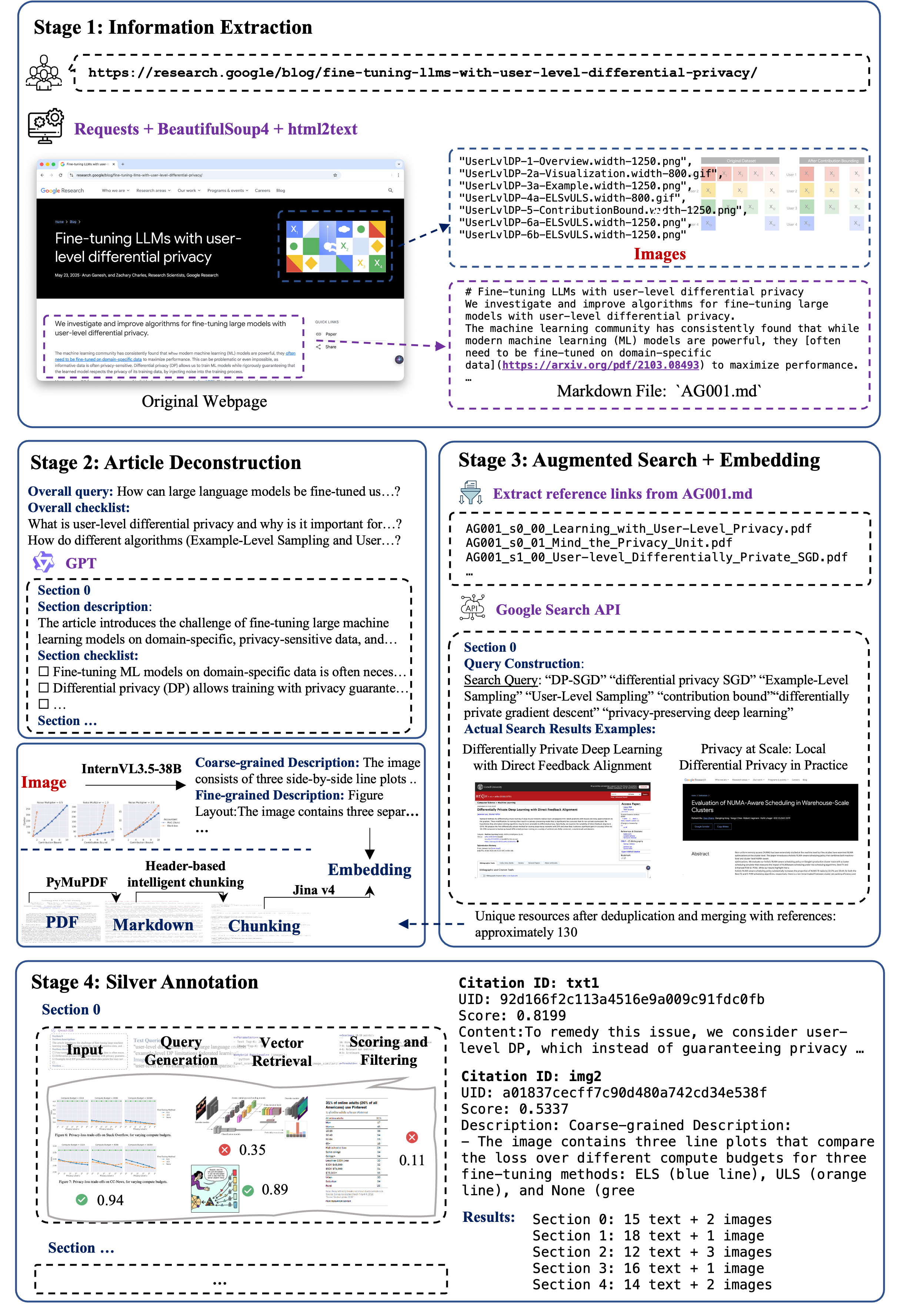}
\end{center}
\clearpage
\begin{center}
    \includegraphics[width=\textwidth]{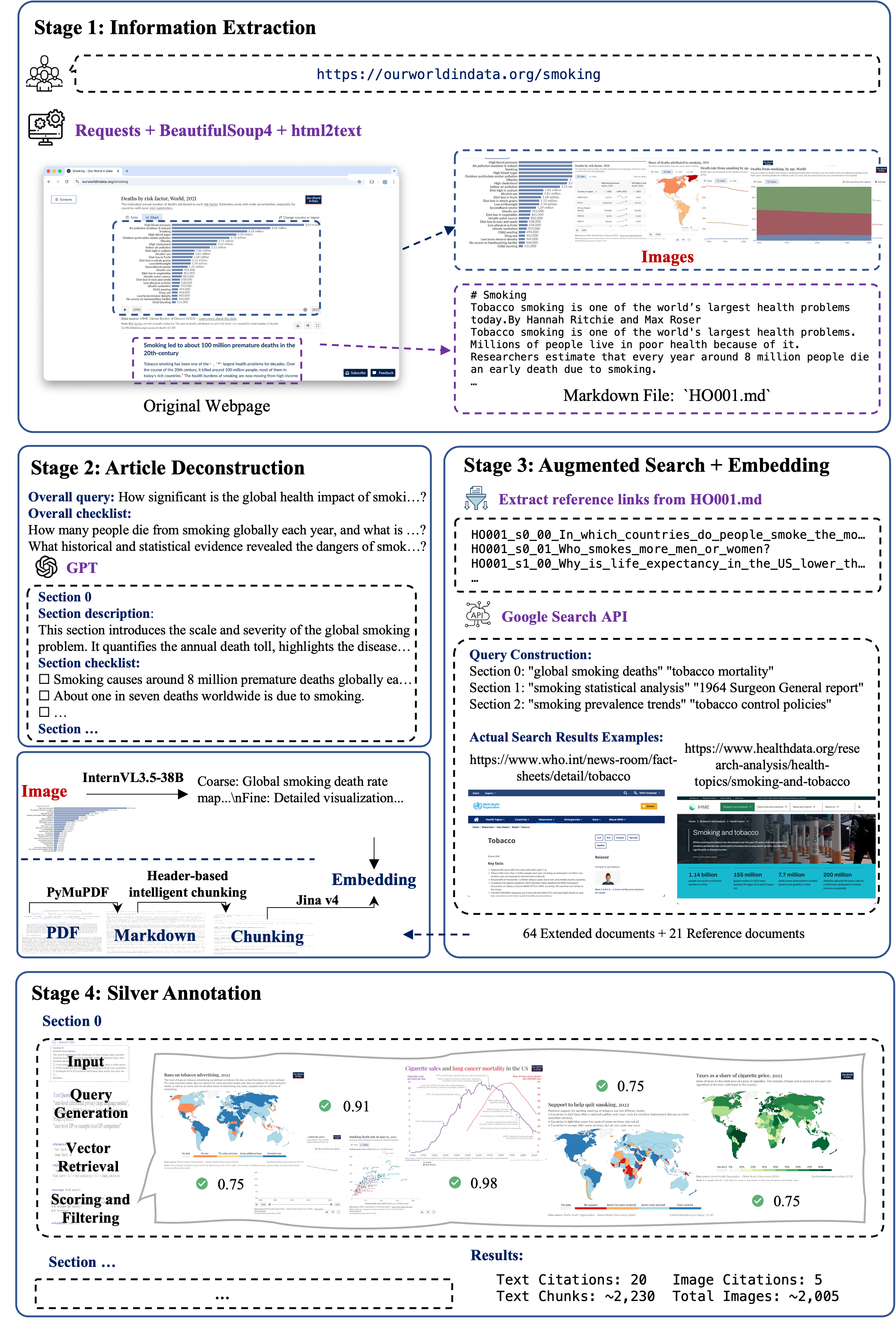}
\end{center}
\clearpage
\begin{center}
    \includegraphics[width=\textwidth]{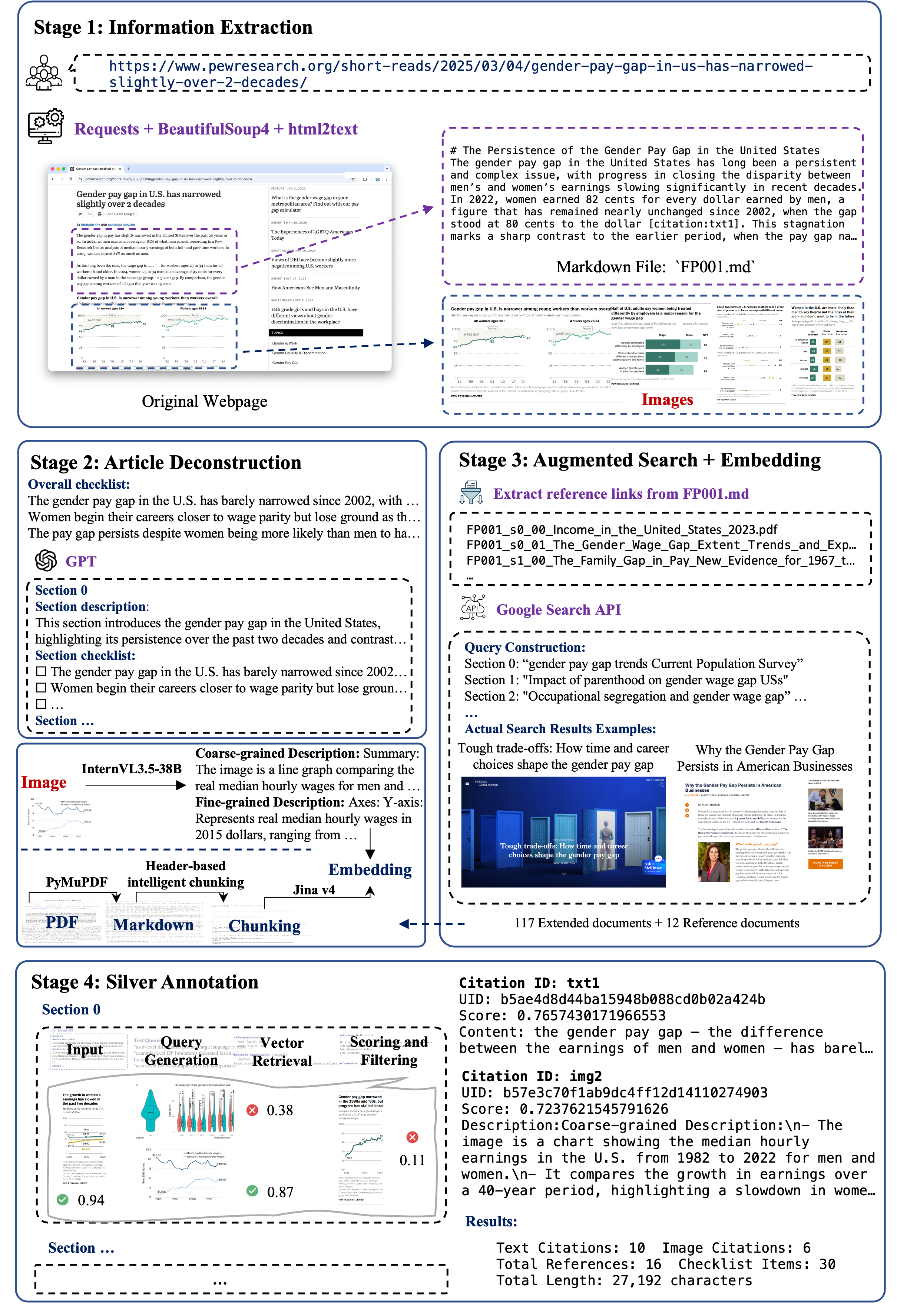}
\end{center}
\clearpage

\section{Experimental Setup Details}
\label{app:exp_details}

\subsection{Model Versions and Licenses}
\label{sub:model_versions}
Our experiments employ the following models: Qwen3-8B, Qwen3-32B, Llama-3.3-70B-Instruct, InternVL3-5-38B, and Jina-Embeddings-v4.

Qwen3-8B, Qwen3-32B and InternVL3-5-38B  are released under the Apache License, Version~2.0. Jina-Embeddings-v4 are released under the Qwen Research License Agreement. Llama-3.3-70B-Instruct is released under the Llama~3.3 Community License Agreement, which is based on the Apache License, Version~2.0.

All models are used in compliance with their respective licenses as released by the original authors, and no restrictions imposed by these licenses are violated in our experimental setting.

The exact model versions, and revision identifiers are summarized in Table~\ref{tab:model_versions}.
\begin{table}[ht]
\centering 
\tiny
\begin{tabular}{@{}p{0.35\columnwidth}p{0.65\columnwidth}@{}}
\textbf{Model} & \textbf{Revision (Git commit SHA)} \\
\midrule
\textbf{Qwen3-8B} & \path{b968826d9c46dd6066d109eabc6255188de91218} \\
\textbf{Qwen3-32B} & \path{9216db5781bf21249d130ec9da846c4624c16137} \\
\textbf{Llama-3.3-70B-Instruct} & \path{6f6073b423013f6a7d4d9f39144961bfbfbc386b} \\
\textbf{InternVL3-5-38B} & \path{de99855be3642cd44fe97c9b72d70e5ce2c07f69} \\
\textbf{Jina-Embeddings-v4} & \path{5f4b9cbb80cc95ba44fe6667dfd75710f7db2947} \\
\end{tabular}
\caption{Model versions and Hugging Face Hub revisions used in our experiments.}
\label{tab:model_versions}
\end{table}

\subsection{System Configuration Details}
\label{sub:system_config}

\textit{\textbf{\quad Checklist-Guided Experimental Control.}}
To ensure controlled and reproducible experiments for open-ended multimodal long-form generation, all reported results condition on a fixed, pre-defined checklist for each task.
These checklists are derived during benchmark construction and specify section-level requirements.
Accordingly, the planner component in \mname is disabled in all experiments, and models are compared solely on their ability to retrieve, filter, and synthesize multimodal evidence under identical structural constraints.
This design choice isolates the effects of agentic orchestration and evidence utilization from variability introduced by planning.

\textit{\textbf{System Configuration of \mname.}}
The Filter module is fixed and kept non-trainable across all experiments to ensure consistent relevance assessment.
Specifically, the Filter employs \texttt{qwen-plus-latest} via API for textual relevance scoring and a locally deployed InternVL3.5-38B model for visual relevance estimation.
The Searcher and Reporter agents share the same backbone model within each experiment (e.g., Qwen3-8B or Qwen3-32B) and are jointly trained when post-training is applied.
This configuration ensures that performance differences arise from agentic coordination and training effects, rather than changes in filtering criteria.

\textit{\textbf{Training Scope and Consistency.}}
Post-training is applied jointly to the Searcher and Reporter agents using curated agentic trajectories.
The Filter module remains unchanged throughout training and evaluation.
This separation guarantees that comparisons across trained and untrained variants, as well as ablations involving the Filter module, are conducted under identical system configurations.

\textit{\textbf{Na\"ive RAG Baselines.}}
Na\"ive RAG baselines follow a structured but non-agentic pipeline.
Given the same section-level checklists and descriptions, the baseline performs parallel single-round retrieval for each section, retrieving top-$10$ text chunks and top-$5$ images per section.
All retrieved multimodal evidence is concatenated and provided to the backbone model in a single generation pass conditioned on the full checklist, without iterative retrieval, relevance-aware filtering, or incremental synthesis.

\subsection{Framework System Prompts}
\label{sub:framework_prompts}
The following system prompts govern the behavior of the agents in \mname.

\onecolumn
    \begin{promptbox}[title={Prompt Template: Outline Generation}]
\begin{lstlisting}[style=promptstyle]
You are an expert at generating article outlines.

Given an overall query/topic and requirements checklist, generate a detailed outline for a long-form article.

**Overall Query/Topic:**
{overall_query}

**Overall Requirements Checklist:**
{overall_checklist}

**Task:**
Generate a structured outline with 3-7 sections. For each section, provide:

1. **section_description**: A DETAILED description (2-4 sentences) that clearly explains:
   - What this section is about
   - What specific aspects will be covered
   - How it contributes to the overall article
   - What the reader should learn from this section

   **IMPORTANT**: DO NOT just write a title or short phrase. Write a comprehensive description that provides enough context for a writer to understand what needs to be written.

2. **sectional_checklist**: A list of 3-5 specific, actionable requirements for this section

**Output Format (JSON):**
```json
{
  "outline": [
    {
      "section_description": "This section provides a comprehensive introduction to the transformer architecture...",
      "sectional_checklist": [
        "Define what transformers are and their primary purpose in deep learning",
        "Explain the historical context and limitations of previous sequence models (RNNs, LSTMs)",
        "Preview the key innovations that make transformers effective",
        "Establish the importance and widespread adoption of transformer architectures"
      ]
    },
    ...
  ]
}
```
Key Guidelines:
- Each section_description should be 2-4 sentences providing rich context
- Avoid single-sentence or title-only descriptions
- Be specific about what content will be covered and why
- Think about what information a writer would need to create high-quality content

Generate the outline now: 
\end{lstlisting} 
\end{promptbox}

    \clearpage
    \begin{promptbox}[title={Prompt Template: Query Generation}]
\begin{lstlisting}[style=promptstyle]
You are an expert at generating search queries for multimodal information retrieval.

Given the article context and current section requirements, generate targeted search queries to find relevant text passages and images.

**Overall Article Query/Topic:**
{overall_query}

**Overall Requirements:**
{overall_checklist}

**Previous Sections Summary:**
{previous_sections_summary}

**Current Section Description:**
{section_description}

**Current Section Requirements:**
{sectional_checklist}

**Task:**
Generate search queries that will help retrieve relevant information for writing this section.
Generate TWO types of queries:
1. **text_queries**: For finding related text passages, explanations, or documentation (3-5 queries)
2. **image_queries**: For finding relevant diagrams, charts, figures, or visualizations (2-4 queries)

**Output Format (JSON):**
```json
{{
  "text_queries": [
    "query 1 for text retrieval",
    "query 2 for text retrieval",
    "query 3 for text retrieval"
  ],
  "image_queries": [
    "query 1 for image retrieval",
    "query 2 for image retrieval"
  ]
}}
```

Generate the queries now:
\end{lstlisting} 
\end{promptbox}

    \clearpage
    \begin{promptbox}[title={Prompt Template: Section Writing}]
\begin{lstlisting}[style=promptstyle]
You are an expert writer skilled at creating coherent, well-structured long-form content with proper source citations.

# Task
Write a specific section of a larger article based on the provided requirements, context, and retrieved materials.

# Overall Article Context
**Article Topic:** {overall_query}

**Overall Requirements:**
{overall_checklist}

# Current Section Information
**Section Position:** {position_info}
**Section Description:** {section_description}

**Section Requirements (must address all):**
{sectional_checklist}

# Context from Previous Sections
**Summary of previous sections:**
{previous_sections_summary}

**End of previous section (for smooth transition):**
{previous_section_tail}

# Retrieved Materials
You have access to the following sources for citation:
{retrieved_materials}

# CRITICAL Citation Rules
You MUST follow these citation rules strictly:
1. **Text Citation Format:**
   - When using information from a text source, cite it as: `[citation:txt1]`, `[citation:txt2]`, etc.
   - Example: "Recent studies show significant progress in AI safety[citation:txt3]."
2. **Image Citation Format:**
   - When you want to insert an image/figure, use: `![](citation:img1)`, `![](citation:img2)`, etc.
   - Example: "The architecture is shown below:\n![](citation:img5)\nAs illustrated, the system..."
   - ONLY use this format for sources explicitly marked as "Image"
3. **Citation Placement:**
   - Place text citations at the end of the sentence or claim
   - Place image citations on their own line where the visual should appear
   - You can cite the same source multiple times if needed
4. **Important Distinctions:**
   - Text sources -> use `[citation:txt1]`, `[citation:txt2]`, etc.
   - Image sources -> use `![](citation:img1)`, `![](citation:img2)`, etc.

# Position-Aware Writing Instructions

**Is this the FIRST section?** {is_first_section}
**Is this the LAST section?** {is_last_section}

## If FIRST section (is_first_section = yes):
- Start with an engaging introduction
- No need to reference previous content (there is none)
- Set the stage for the article
- DO NOT conclude the entire article

## If MIDDLE section (is_first_section = no, is_last_section = no):
- Begin with a smooth transition from the previous section
- Focus ONLY on this section's specific requirements
- **CRITICAL: DO NOT use concluding phrases like:**
  - "In conclusion"
  - "To summarize"
  - "Overall, this article"
  - "We have discussed"
  - "In this article, we covered"
- Continue the narrative without wrapping up the entire article
- End in a way that allows the next section to continue naturally

## If LAST section (is_last_section = yes):
- Provide appropriate transitions from previous content
- You MAY synthesize and conclude the entire article
- Wrap up all major points discussed in the article

# Content Requirements
1. **Address all sectional requirements** in the checklist above
2. **Maintain coherence** with previous sections using the provided context
3. **Use retrieved materials** as the foundation of your content
4. **Cite sources properly** using the formats specified above
5. **Write naturally** - your content should flow smoothly while integrating citations
6. **Appropriate length** - typically 300-800 words depending on the complexity of requirements

# Markdown Formatting Requirements
**CRITICAL: You MUST use proper Markdown formatting for structure:**

1. **Section Title (Required):**
   - Start your content with a section title using `# Title`
   - The title should reflect the section description

2. **Subsections (If Needed):**
   - Use `## Subsection Title` for major subsections
   - Use `### Subsubsection Title` for deeper nesting
   - Use appropriate heading levels to create clear hierarchy

3. **Example Structure:**
```markdown
# Introduction to Machine Learning

Machine learning has revolutionized...[citation:txt1]

## Supervised Learning
Supervised learning approaches...[citation:txt2]

### Classification
Classification tasks involve...[citation:txt3]

## Unsupervised Learning
Unlike supervised methods...[citation:txt5]
```

# Output Format
Provide ONLY the section content with proper Markdown formatting and citations. Do NOT include:
- Meta-commentary about what you're doing
- Explanations of your writing process

**Remember to start with a section title using `#`!**
Begin writing the section now:
\end{lstlisting} 
\end{promptbox}

    \clearpage
    \begin{promptbox}[title={Prompt Template: Image Filter}]
\begin{lstlisting}[style=promptstyle]
You are an expert at evaluating image relevance for academic/technical content.

# Context
**Current Section Description:** {section_description}

**Section Requirements:**
{sectional_checklist}

**Image Description:** {image_description}

# Question
Based on the section requirements above, is this image relevant and useful for this section?

Consider:
1. Does the image visually illustrate concepts mentioned in the requirements?
2. Is it a diagram, chart, figure, or visualization that adds value?
3. Does it relate to the section topic?

# Response Format
Answer with ONLY ONE WORD:
- "YES" if the image is relevant and should be kept
- "NO" if the image is not relevant

Your answer:
\end{lstlisting} 
\end{promptbox}

    \begin{promptbox}[title={Prompt Template: Text Filter}]
\begin{lstlisting}[style=promptstyle]
You are an efficient information filtering assistant.

# Task
Determine which of the provided information sources are highly relevant to the current section requirements.

# Context
**Current Section Description:** {section_description}

**Section Requirements:**
{sectional_checklist}

**Information Sources to Filter:**
{sources_list}

# Instructions
1. Read each numbered source carefully (txt1, txt2, txt3, etc.)
2. Determine if it provides valuable information for the section requirements
3. Return the numbers of ALL useful sources

# Output Format
Return ONLY a list of numbers (comma-separated or Python list format):
- Correct example 1: 1, 3, 5, 8
- Correct example 2: [1, 3, 5, 8]
- If none are useful: []

Note: The sources are labeled as txt1, txt2, etc., but you should return just the numbers (1, 2, 3...).

Your response:
\end{lstlisting} 
\end{promptbox}

\twocolumn

\section{Training Implementation}
\label{app:training_details}

\subsection{Supervised Fine-Tuning (SFT) Parameters}
\label{sub:sft_params}
SFT was performed on 8$\times$H100 GPUs. The exact training configuration is summarized in Table~\ref{tab:sft_params}.
\begin{table}[ht]
    \centering
    \small
    \begin{tabular}{l|c}
        \toprule
        \textbf{Parameter} & \textbf{Value} \\
        \midrule
        Training Objective & Next-token prediction \\
        Optimizer & AdamW ($\beta_1$=0.9, $\beta_2$=0.95) \\
        Learning Rate & 1e-4 \\
        LR Scheduler & Cosine decay \\
        Warmup Ratio & 0.05 \\
        Global Batch Size & 64 \\
        Per-device Batch Size & 1 \\
        Grad. Accumulation Steps & 8 \\
        Weight Decay & 0.1 \\
        Max Sequence Length & 32768 \\
        Epochs & 1 \\
        LoRA Rank ($r$) & 16 \\
        LoRA Alpha ($\alpha$) & 32 \\
        LoRA Dropout & 0.05 \\
        Target Modules & all-linear \\
        Precision & bfloat16 \\
        DeepSpeed & ZeRO-3 \\
        Gradient Checkpointing & \checkmark \\
        \bottomrule
    \end{tabular}
    \caption{Supervised Fine-Tuning (SFT) hyperparameters.}
    \label{tab:sft_params}
\end{table}

Taking Qwen3-32B as an example, the base model contains approximately 32.9B parameters, of which only 134.2M (0.41\%) are trainable through LoRA adaptation. Training on the SFT dataset of 8k samples completed in approximately 1 hour 53 minutes, achieving a final training loss of 1.20 and token accuracy of 66.8\%. Peak GPU memory usage reached 31.5 GiB per device.

\subsection{DPO Configuration}
\label{sub:dpo_config}

DPO training was conducted using the same 8$\times$H100 GPU configuration. Table~\ref{tab:dpo_params} details the hyperparameters used. Specifically, for each input (task and initial state), we form a pair consisting of a \emph{chosen} trajectory $y_w$ and a \emph{rejected} trajectory $y_l$. The \emph{chosen} trajectories are sampled from our collected dataset of 8K human-curated traces, representing successful or higher-quality executions. In contrast, the \emph{rejected} trajectories are generated autonomously by the model under the same input conditions, and are treated as lower-quality alternatives during pairing.
\begin{table}[ht]
    \centering
    \small
    \begin{tabular}{l|c}
        \toprule
        \textbf{Parameter} & \textbf{Value} \\
        \midrule
        Loss Type & DPO (sigmoid) \\
        Beta ($\beta$) & 0.1 \\
        Learning Rate & 5e-6 \\
        LR Scheduler & Cosine decay \\
        Warmup Ratio & 0.05 \\
        Global Batch Size & 64 \\
        Per-device Batch Size & 1 \\
        Grad. Accumulation Steps & 8 \\
        Weight Decay & 0.1 \\
        Max Sequence Length & 20480 \\
        Epochs & 1 \\
        Optimizer & AdamW ($\beta_1$=0.9, $\beta_2$=0.95) \\
        LoRA Rank ($r$) & 16 \\
        LoRA Alpha ($\alpha$) & 32 \\
        LoRA Dropout & 0.05 \\
        Target Modules & all-linear \\
        Precision & bfloat16 \\
        DeepSpeed & ZeRO-3 \\
        Gradient Checkpointing & \checkmark \\
        \bottomrule
    \end{tabular}
    \caption{Direct Preference Optimization (DPO) configuration.}
    \label{tab:dpo_params}
\end{table}

For Qwen3-32B, the DPO phase utilized 8k preference pairs with longer average sequence lengths (6,946 vs. 3,863 tokens in SFT), resulting in higher computational demands. Training completed in approximately 3 hours 9 minutes with a final loss of 0.024 and reward accuracy converging to 100\%. Due to the paired comparison nature of DPO, peak memory usage increased to 73.2 GiB per device.

\newpage
\section{Evaluation Details}
\label{app:eval_details}

\subsection{Scoring Rubrics}
\label{sub:scoring_details}

We assess report quality across three hierarchical dimensions, each capturing distinct aspects of generation quality:

\textit{\textbf{Section Anchor Quality}} evaluates structural adherence to planning constraints (\S\ref{subsec:framework}). This includes \textbf{(1) Description Adherence $S_{\text{Desc}}$:} semantic alignment between generated content and the planner's intent definition, and \textbf{(2) Checklist Satisfaction $S_{\text{Check}}$:} coverage of specific requirements from the checklist, where the judge provides reasoning-backed scores for each item and penalizes superficial mentions or missing required facts.

\textit{\textbf{Section Content Quality}} assesses multimodal grounding through four complementary aspects: \textbf{(1) Richness $S_{\text{Rich}}$:} whether the quantity and diversity of visual aids (diagrams, charts, photos, etc.) match the informational density and are well-distributed throughout the section; \textbf{(2) Image-Text Coherence $S_{\text{i-Coh}}$:} whether images directly relate to and support the immediate textual context, with visual elements aligned to textual explanations and each image adding meaningful information rather than being tangential; \textbf{(3) Placement $S_{\text{Place}}$:} whether images are positioned at logical points in the text flow with appropriate contextual setup before/after each image, appearing when relevant concepts are discussed (not too early or late); and \textbf{(4) Visual Clarity $S_{\text{Clarity}}$:} whether images are clear and understandable with discernible key elements, appropriate for academic/technical contexts, and maintain consistent quality. We implement adaptive scoring: sections describing architectures, processes, or experimental results are penalized for missing images, while purely conceptual or abstract theoretical sections without images receive high scores.

\textit{\textbf{Full Report Quality}} performs holistic evaluation of narrative integrity through \textbf{(1) Flow Coherence $S_{\text{Coh}}$:} logical transitions and narrative smoothness between sections; \textbf{(2) Fluency $S_{\text{Flu}}$:} grammatical correctness and professional tone maintenance; \textbf{(3) Repetition Absence $S_{\text{Rep}}$:} absence of redundant information loops or repetitive phrasing across sections (higher scores indicate less repetition); and \textbf{(4) Termination Quality $S_{\text{Term}}$:} absence of premature concluding statements (e.g., "In summary") in non-terminal sections.

\subsection{Relative Quality Score Calculation}
\label{sub:win_rate_calc}

In Table~\ref{tab:deepreporter_generation}, we report the \textit{Relative Quality Score} to provide a normalized, bias-resistant assessment of multimodal generation quality. This metric is designed to address two fundamental challenges in LLM evaluation: positional bias in comparative judgments and uncalibrated scoring across varying task difficulties.

\textit{\textbf{Challenge 1: Positional and Length Bias in Pairwise Comparisons.}}
Standard pairwise evaluation protocols (\textit{e.g., "Which is better: Output A or Output B?"}) are susceptible to systematic biases where judges favor the first-presented response or longer responses regardless of actual quality \citep{wang2024large, tripathi2025pairwise}. These artifacts can dominate quality signals, particularly in long-form generation tasks where outputs vary significantly in structure and length.

To eliminate these confounding factors, we adopt an independent pointwise evaluation protocol. Each model's output is assessed solely based on its intrinsic quality against the rubrics defined in \S\ref{sub:scoring_details}, without direct comparison to other models' outputs. This design ensures that scores reflect content quality rather than presentation order or verbosity.

\textit{\textbf{Challenge 2: Uncalibrated Scoring Across Task Difficulties.}}
Raw scalar scores (\textit{e.g., 7.5 vs. 7.8}) are inherently difficult to interpret across tasks of varying complexity. A score of 8.0 on a simple descriptive task does not carry the same semantic weight as 8.0 on a highly technical analytical report.

Following robust reference-guided benchmarks~\cite{dubois2024length,lin2024wildbench}, we normalize model performance against the Expert-Refined Silver Reference $\mathcal{R}_{\text{ref}}$ (constructed in \S\ref{sec:benchmark}). This reference-relative approach transforms absolute scores into a calibrated measure of expert-level consistency.

\textit{\textbf{Calculation 1: Raw Score Acquisition.}}
For each sub-metric $m$ defined in \S\ref{sub:scoring_details}, we first obtain a raw scalar score $S_{\text{model}, i}^{(m,raw)} \in [0, 10]$ for task $i$. We employ specialized judges for different dimensions to ensure evaluation accuracy: \textbf{(1) For Section Anchor and Full Report dimensions}, we employ GPT-4.1~\cite{achiam2023gpt} as the judge, assigning continuous scores (0--10) across four performance tiers: Excellent (8--10), Good (6--8), Average (4--6), and Poor (0--4). The judge receives the section description, checklist requirements, and generated content to assess structural adherence and narrative quality. \textbf{(2) For Section Content dimension}, we use \href{https://huggingface.co/OpenGVLab/InternVL3_5-38B}{InternVL3-5-38B}\footnote{Using \href{https://github.com/vllm-project/vllm}{vLLM} to deploy for efficient and stable inference.}~\cite{wang2025internvl3} to evaluate multimodal grounding quality on the same 0--10 scale. The judge examines the actual rendered images alongside the section content to assess visual element quality, placement, and coherence.

\textit{\textbf{Calculation 2: Relative Score Scaling.}}
Raw scalar scores are inherently difficult to interpret across tasks of varying complexity. To calibrate these scores, we normalize model performance against the Expert-Refined Silver Reference $\mathcal{R}_{\text{ref}}$ (constructed in \S\ref{sec:benchmark}).

We define the \textit{Relative Quality Score} ($\text{Score}_m$) for sub-metric $m$ as the percentage of tasks where the model meets or exceeds expert-level performance:
\begin{equation}
\text{Score}_m = \frac{1}{N} \sum_{i=1}^{N} \mathbb{I}\left( S_{\text{model}, i}^{(m,raw)} \geq S_{\text{ref}, i}^{(m,raw)} \right) \times 100
\end{equation}
where $N=247$ is the total number of tasks, and $\mathbb{I}(\cdot)$ is the indicator function. This binary comparison eliminates potential positional biases found in direct pairwise preferences and focuses solely on whether the model reaches the expert baseline.

\textit{\textbf{Calculation 3: Hierarchical Aggregation.}}
Finally, we aggregate the fine-grained metric scores into dimension-level scores and an overall quality score. This allows us to observe whether a model consistently meets standards on specific aspects rather than allowing strengths in one area to mask weaknesses in another.

We compute the dimension-level scores as the arithmetic mean of their constituent sub-metrics:
\begin{align}
S_{\text{Anchor}} &= (S_{\text{Desc}} + S_{\text{Check}})/2 \\
S_{\text{Content}} &= (S_{\text{Rich}} + S_{\text{iCoh}} + S_{\text{Pla.}} + S_{\text{Clar}})/4 \\
S_{\text{Report}} &= (S_{\text{Coh}} + S_{\text{Flu}} + S_{\text{Rep}} + S_{\text{Term}})/4
\end{align}

The \textit{Overall Relative Quality Score} reported in our main results is the mean of these three high-level dimensions:
\begin{equation}
\text{Overall} = \frac{S_{\text{Anchor}} + S_{\text{Content}} + S_{\text{Report}}}{3}
\end{equation}
This hierarchical formulation provides a holistic view of the model's capability to generate reports that are structurally sound, multimodally grounded, and narratively coherent.

\subsection{Human-AI Agreement Study}
\label{sub:human_agreement}
To validate the reliability of the automated evaluation metrics, we conducted a blind
human--AI agreement study.
Human annotators independently scored each report across multiple evaluation dimensions
using a 5-point Likert scale (1 = very poor, 5 = excellent).
For the same set of reports, the automated evaluator (GPT-4) produced continuous scores
on a 0--10 scale following a strict evaluation rubric.

To enable a fair comparison between human and automated judgments,
human scores were linearly rescaled to the same 0--10 range prior to analysis.
Specifically, human ratings of 1--5 were mapped to 2, 4, 6, 8, and 10, respectively.
Agreement was measured using the Pearson correlation coefficient between the averaged
human scores and the GPT-4 scores for each evaluation dimension.

We randomly sampled 30 reports from the evaluation set and asked three PhD-level domain
experts to provide independent annotations.
Human--AI agreement was computed separately for all evaluation dimensions, including
\textit{Section}-level metrics (Description, and Check),
\textit{Article}-level metrics (Coherence, Fluency, Repetition, and Terminology),
\textit{Multimodal}-level metrics (Richness, Coherence, Placement, and Clarity), and \textit{Overall}.

\subsection{Evaluation Costs}
\label{sub:eval_cost}
Evaluating 247 papers with LLM- and VLM-based automated metrics resulted in an approximate total
API cost of \$15.
Text-only evaluation accounted for roughly \$10, while multimodal (image-based)
evaluation contributed an additional \$5.

\onecolumn
    \clearpage
    \begin{promptbox}[title={Prompt Template: Article Evaluation and Scoring}]
\begin{lstlisting}[style=promptstyle]
You are a strict, meticulous, and objective expert evaluator for long-form generated articles. Your task is to evaluate both individual sections and the overall article quality.

## Core Scoring Rules (Apply to ALL scores)
1. **Use a scale of 0-10 (continuous values)**: Do not cluster scores around 8-10. 
   - **8-10 points**: Excellent/outstanding performance. Fully meets or exceeds requirements.
   - **6-8 points**: Good performance. Largely meets requirements with notable strengths.
   - **4-6 points**: Average performance. Basically meets requirements, neither good nor bad.
   - **2-4 points**: Poor performance. Minimally meets requirements.
   - **0-2 points**: Very poor performance. Almost completely failed or missing.
2. **Be Harsh**: Default to a lower score if you are unsure. Penalize superficial content heavily.
## Article Overview
**Query**: {query}
**Overall Checklist**: {overall_checklist}
**Number of Sections**: {num_sections}

## Evaluation Tasks

### Part 1: Section-Level Evaluation
For each section, you will evaluate:
1. **Description Completion Score (0-10)**: How well does the generated content match the intended section description?
2. **Checklist Completion**: For each item in the sectional checklist, evaluate the quality of the coverage with a brief explanation.

### Part 2: Article-Level Evaluation
For the entire article, evaluate the following dimensions (each scored 0-10):
1. **Coherence**: How well do the sections connect with each other? Are there smooth transitions? Is there a logical flow?
2. **Fluency**: Is the writing clear, natural, and easy to read? Are sentences well-constructed?
3. **Repetition**: Are there unnecessary repetitions across sections?
4. **Termination**: Are there inappropriate concluding statements in non-conclusion sections? (Lower score = more inappropriate conclusions)

## Sections Data
{sections_data}

## Output Format
Respond with a valid JSON object (no markdown code blocks, just raw JSON) with the structure:
{{
  "section_evaluations": [
    {{
      "section_index": 0,
      "description_completion_score": Continuous score 0-10,
      "description_completion_reasoning": "Brief explanation of the score",
      "checklist_evaluations": [
        {{
          "checklist_item": "The exact checklist item text",
          "score": Continuous score 0-10,
          "reasoning": "Brief explanation"
        }}
      ]
    }}
  ],
  "article_evaluation": {{
    "coherence_score": Continuous score 0-10,
    "coherence_reasoning": "Brief explanation",
    "fluency_score": Continuous score 0-10,
    "fluency_reasoning": "Brief explanation",
    "repetition_score": Continuous score 0-10,
    "repetition_reasoning": "Brief explanation (note: higher = less repetition)",
    "termination_score": Continuous score 0-10,
    "termination_reasoning": "Brief explanation (note: higher = fewer inappropriate conclusions)"
  }}
}}
\end{lstlisting} 
\end{promptbox}

    \clearpage
    \begin{promptbox}[title={Prompt Template: Image Quality Evaluation}]
\begin{lstlisting}[style=promptstyle]
You are a strict, meticulous, and objective expert evaluator for multimodal content in academic writing. Your task is to evaluate the quality of image usage in a section of an article.

## Core Scoring Rules (Apply to ALL scores)
1. **Use a scale of 0-10 (continuous values)**: Do not cluster scores around 8-10.
   - **8-10 points**: Excellent/outstanding performance. Fully meets or exceeds requirements.
   - **6-8 points**: Good performance. Largely meets requirements with notable strengths.
   - **4-6 points**: Average performance. Basically meets requirements, neither good nor bad.
   - **2-4 points**: Poor performance. Minimally meets requirements.
   - **0-2 points**: Very poor performance. Almost completely failed or missing.
2. **Be Harsh**: Score lower if unsure. Penalize poor images heavily.
3. **Evaluate All Dimensions Independently**: Dimensions assess image usage quality.
4. **IMPORTANT - Always Return Numeric Scores**: Never use "N/A", "None", or text descriptions as scores. Always provide a numeric value (0-10) based on the evaluation criteria below.

## Section Context
**Section Description**: {section_description}

**Section Content (with image placeholders)**:
{section_content}

**Number of Images in Section**: {num_images}

## Images to Evaluate
{images_info}
**CRITICAL INSTRUCTION**:
- You MUST carefully examine the ACTUAL IMAGES provided below (not text descriptions)
- Base your evaluation on what you directly observe in the visual content
- Analyze the visual elements, layout, clarity, and relevance that you see
- Do not rely on or assume content based on text descriptions or expectations

## Evaluation Task
Evaluate the section's image usage based on the following dimensions (each scored 0-10):

### 1. Image Richness (richness_score)
Evaluate whether the quantity and variety of images are appropriate:
- Is the number of images suitable for the section length and complexity?
- Are images distributed well throughout the section (not clustered)?
- Does the variety of images (diagrams, charts, photos, etc.) match the content needs?

**Special Case - If there are NO images (num_images = 0):**
- Analyze the section content to determine if images are necessary
- **If images are NOT needed** (e.g., conceptual discussion, definitions, abstract theory):
  - Score: 7-10 points (no images needed, appropriate decision)
- **If images ARE needed** (e.g., describing architectures, processes, experimental results):
  - Score: 0-3 points (missing essential visual aids)
- **Uncertain cases** (e.g., could benefit from images but not strictly necessary):
  - Score: 4-6 points (adequate but could be improved)

### 2. Image-Text Coherence (coherence_score)
Evaluate how well images relate to and support the text content:
- Do all images directly relate to and support the text content?
- Are the visual elements aligned with what the text is explaining?
- Does each image add meaningful information that complements the text?
- Are there any irrelevant or tangentially related images?

**Special Case - If there are NO images:**
- If images are not needed: Score 8-10 (no coherence issues, appropriate)
- If images are needed: Score 0-3 (no support for text, poor coherence)
- If uncertain: Score 5 (neutral, neither good nor bad)

### 3. Placement & Integration (placement_score)
Evaluate the positioning and integration of images within the text flow:
- Are images placed at logical points in the text flow?
- Does the text provide proper context before/after each image?
- Are transitions between text and images smooth and natural?
- Do images appear when relevant concepts are discussed (not too early or late)?

**Special Case - If there are NO images:**
- If images are not needed: Score 8-10 (no placement issues, appropriate)
- If images are needed: Score 0-3 (no integration, missing placement)
- If uncertain: Score 5 (neutral)

### 4. Visual Quality & Clarity (clarity_score)
Evaluate the quality and understandability of the images themselves:
- Are the images clear and easy to understand?
- Are key elements in each image visible and discernible?
- Are images appropriate for the academic/technical context?
- Do images maintain consistent quality and style?

**Special Case - If there are NO images:**
- If images are not needed: Score 8-10 (no quality issues, appropriate)
- If images are needed: Score 0-3 (no visual quality, missing)
- If uncertain: Score 5 (neutral)

## Content Types That Typically NEED Images

Consider the following content types as typically requiring visual aids:
1. **System/Model Architecture**: Diagrams showing components and connections
2. **Algorithms/Methods**: Flowcharts or pseudocode visualizations
3. **Experimental Setup**: Photos or diagrams of equipment/environments
4. **Results/Data**: Charts, graphs, tables showing findings
5. **Processes/Workflows**: Step-by-step visual representations
6. **Comparisons**: Side-by-side visual comparisons
7. **Examples/Case Studies**: Concrete visual examples

## Content Types That May NOT Need Images

1. **Abstract Definitions**: Pure conceptual explanations
2. **Literature Review**: Discussion of related work (unless comparing approaches)
3. **Theoretical Background**: Mathematical proofs, theoretical foundations
4. **Introductory Text**: Background context, motivation
5. **Conclusions**: Summary statements, future work discussions

## Output Format
You MUST respond with ONLY a valid JSON object. Follow this EXACT structure:
{{
  "richness_score": <float between 0-10>,
  "richness_reasoning": "<your explanation here>",
  "coherence_score": <float between 0-10>,
  "coherence_reasoning": "<your explanation here>",
  "placement_score": <float between 0-10>,
  "placement_reasoning": "<your explanation here>",
  "clarity_score": <float between 0-10>,
  "clarity_reasoning": "<your explanation here>"
}}

**CRITICAL OUTPUT REQUIREMENTS - READ CAREFULLY:**
1. **Score Fields Must Contain NUMBERS ONLY** (not text):
   - Replace `<float between 0-10>` with an actual number like 7.5 or 3.2
   - CORRECT: "richness_score": 7.5
   - CORRECT: "richness_score": 3.0
   - WRONG: "richness_score": "This section lacks images"
   - WRONG: "richness_score": "N/A"
   - WRONG: "richness_score": "Since no images are present..."
   - WRONG: "richness_score": "Continuous score 0-10"
2. **Reasoning Fields Must Contain TEXT ONLY**:
   - Replace `<your explanation here>` with your actual explanation
   - CORRECT: "richness_reasoning": "Brief explanation of the score"
   - WRONG: Put explanations in the score field
3. **For sections with NO images**:
   - Still provide NUMERIC scores (not text or "N/A")
   - Low score if images are needed
   - High score if images not needed
   - Put explanation in reasoning field
\end{lstlisting} 
\end{promptbox}

\twocolumn

\section{Additional Experimental Results}
\label{app:add_results}

\begin{table}[t]
\centering
\small
\setlength{\tabcolsep}{3.5pt}
\renewcommand{\arraystretch}{1.05}
\resizebox{\columnwidth}{!}{%
\begin{tabular}{lccccccc}
\toprule
\textbf{Model} & \textbf{Filter} & \textbf{Train} & \textbf{Total} & \textbf{Query} & \textbf{Search} & \textbf{Filter} & \textbf{Write} \\
\midrule
Qwen3-8B       & \cmark & Base & 630 & 10.6 & 30.4 & 541.4 & 47.5 \\
Qwen3-8B       & \xmark & Base & 170 & 18.1 & 75.1 & - & 77.2 \\
Qwen3-8B       & \cmark & SFT & 636 & 12.1 & 31.6 & 530.3 & 61.5 \\
\midrule
Qwen3-32B      & \cmark & Base & 644 & 28.1 & 29.0 & 434.3 & 152.6 \\
Qwen3-32B      & \xmark & Base & 342 & 43.7 & 38.4 & - & 259.6 \\
Qwen3-32B      & \cmark & SFT & 625 & 37.9 & 31.8 & 369.1 & 186.7 \\
\midrule
Llama3.3-70B   & \cmark & SFT & 955 & 194.9 & 38.4 & 245.1 & 477.1 \\
\midrule
GPT-4.1        & \cmark & Base & 636 & 14.6 & 32.9 & 534.5 & 54.1 \\
\bottomrule
\end{tabular}%
}
\caption{End-to-end runtime breakdown (seconds) averaged over 247 tasks. \textbf{Query}: LLM generates search queries. \textbf{Search}: retrieval service latency. \textbf{Filter}: LLM filters retrieved results. \textbf{Write}: LLM generates section content.}
\label{tab:runtime_breakdown}
\end{table}

\subsection{Runtime Breakdown and Cost Analysis}
\label{sub:runtime_analysis}

We provide a detailed runtime breakdown of \mname to complement the main experimental results.
Table~\ref{tab:runtime_breakdown} reports the end-to-end latency averaged over 247 tasks, decomposed into query generation, retrieval, filtering, and section writing.

Overall runtime is dominated by relevance-aware filtering and section-level content generation.
With filtering enabled, the Filter stage alone accounts for a substantial portion of the total latency (e.g., 541.4s for Qwen3-8B and 434.3s for Qwen3-32B), whereas query generation and retrieval incur comparatively minor overhead.
Disabling the Filter significantly reduces total runtime (630s $\rightarrow$ 170s on Qwen3-8B; 644s $\rightarrow$ 342s on Qwen3-32B), but this reduction comes at the cost of degraded multimodal grounding quality, as shown in Table~\ref{tab:deepreporter_generation}.
This confirms that relevance-aware filtering functions as a stabilizing component for long-form multimodal synthesis rather than a lightweight retrieval optimization.

Post-training does not introduce noticeable inference-time overhead.
For both Qwen3-8B and Qwen3-32B, SFT-trained models exhibit similar total runtime and stage-wise distributions compared to their base counterparts (e.g., 644s vs.\ 625s on Qwen3-32B), indicating that performance gains from training stem from improved decision policies rather than increased computation.

As expected, larger backbones incur higher section-writing costs.
For example, Llama3.3-70B spends 477.1s on content generation, substantially more than Qwen-based models, while following a similar distribution across pipeline stages.

From a practical perspective, generating a single multimodal research report with \mname typically requires several minutes of wall-clock time per query.
The dominant cost arises from LLM-based filtering and section writing, while the marginal cost of post-training does not affect inference-time efficiency.

\subsection{Domain-Specific Performance \& Radar Charts}
\label{sub:radar_domain}
\begin{figure}[ht]
    \centering
    \includegraphics[width=\textwidth]{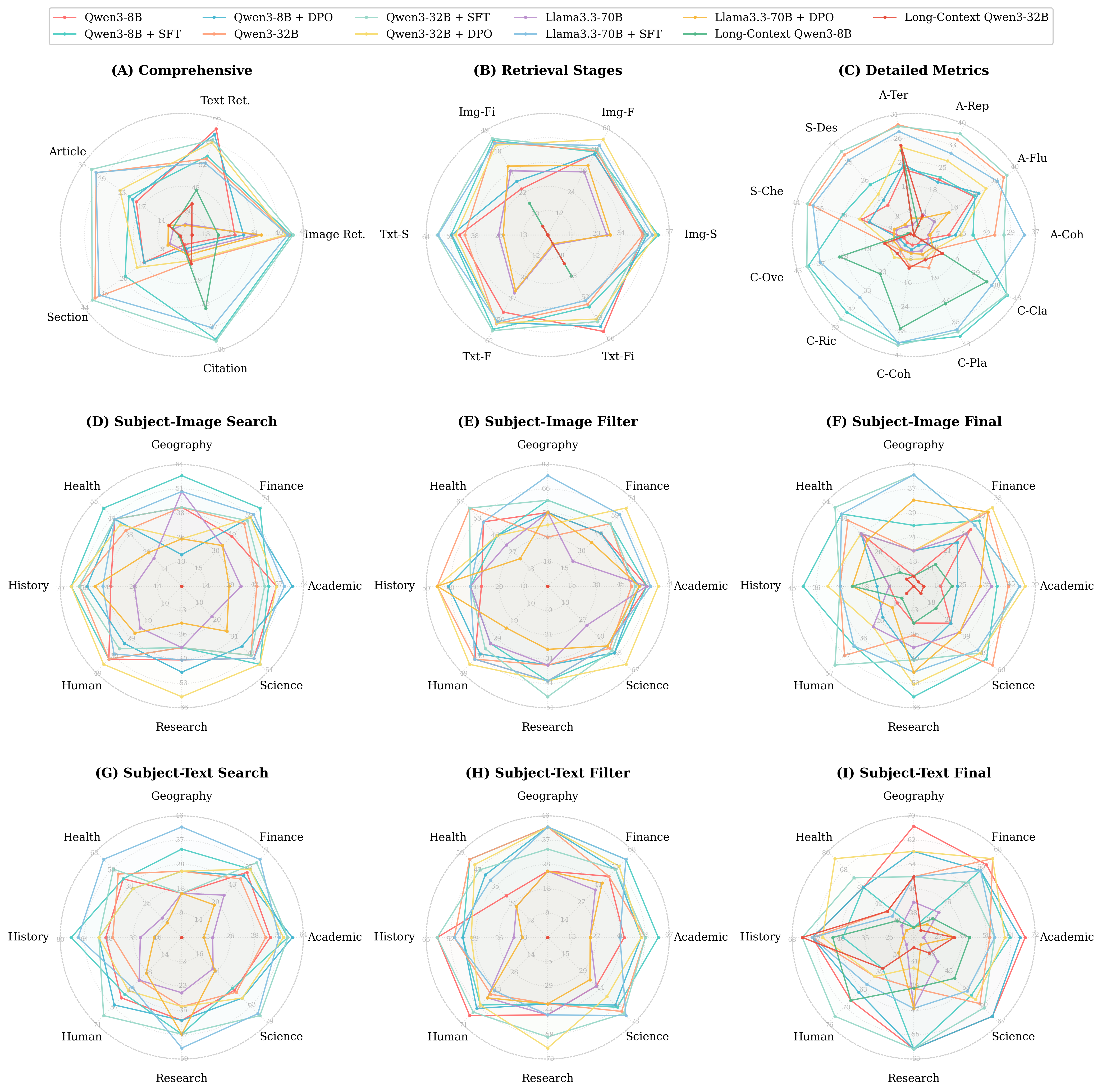}
    \caption{Model performance evaluation. (A-C) Overall comparisons:
  comprehensive metrics, retrieval pipeline stages, and detailed quality indicators. (D-I)
  Subject-specific retrieval performance for image and text modalities across different pipeline
  stages.}
    \label{fig:radar_chart}
\end{figure}

\begin{figure}[ht]
    \centering
    \includegraphics[width=\textwidth]{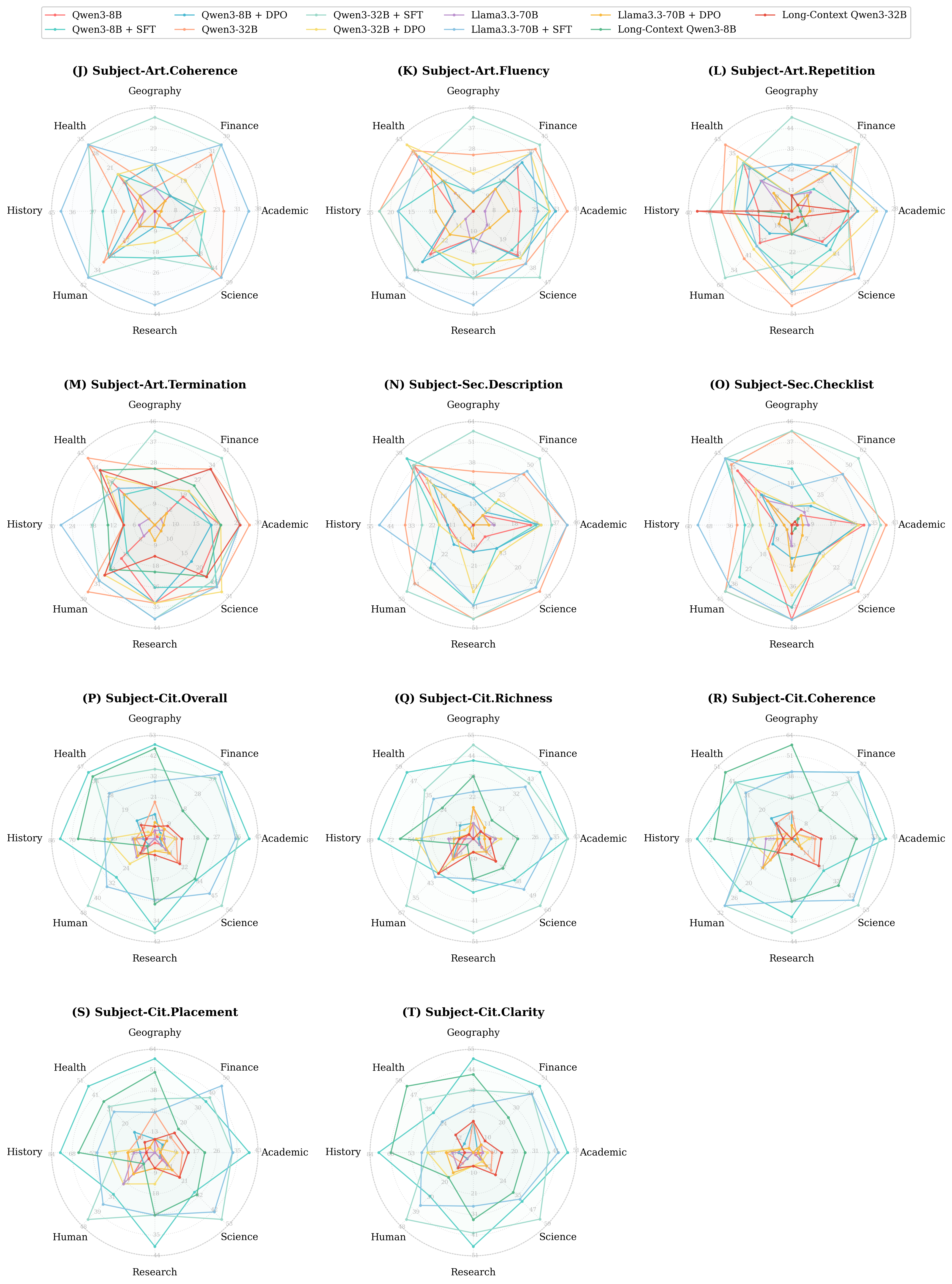}
    \caption{Subject-specific generation quality score. (J-O) Article and section
  quality across six dimensions: coherence, fluency, repetition control, termination, description
  completeness, and checklist adherence. (P-T) Citation quality across five dimensions: overall
  performance, richness, coherence, placement, and clarity. Performance patterns vary significantly
  across academic domains.}
    \label{fig:radar_charts_part2}
\end{figure}

In the main text, we have already reported the results for search and generation in Table~\ref{tab:deepreporter_retrieval} and Table~\ref{tab:deepreporter_generation}. Here, we present finer-grained results by incorporating domain-level (discipline-level) analyses. Specifically, Figure~\ref{fig:radar_chart} first summarizes the overall performance and then breaks down search performance across different disciplines. Figure~\ref{fig:radar_charts_part2} presents the results for the generation task, likewise analyzed by discipline.

Consistent with our earlier findings, in Figure~\ref{fig:radar_chart} (A–C) the models achieve relatively similar performance on retrieval-related metrics (Ret), while SFT yields the best performance in the final results. This trend is also reflected in the intermediate search results across different stages for both image and text evidence. From a domain perspective, model performance varies across disciplines. Moreover, strong performance in a given discipline under text-based retrieval does not necessarily translate to similarly strong performance under image-based retrieval (and vice versa), highlighting the impact of modality differences on domain-level behavior.

Based on Figure~\ref{fig:radar_charts_part2}, we further analyze domain-specific generation performance. When examining generation performance at different levels of granularity, we observe that as the granularity becomes finer, the consistency of different models across evaluation metrics generally improves.
More concretely, at the section and citation level, a certain degree of cross-metric consistency emerges within the same model across different disciplines. For example, for the long-context Qwen3-8B model, performance in the Human domain is consistently lower than in other disciplines across multiple metrics, whereas performance in the Health domain is comparatively strong in most cases.

A similar pattern can be observed for Qwen3-32B + SFT, where performance in the History domain is consistently weaker than in other domains across different generation metrics.

\onecolumn

\onecolumn
\twocolumn

\subsection{Analysis of DPO Underperformance}
\label{sub:sft_dpo}
Table~\ref{tab:deepreporter_retrieval} analyzes the retrieval pipeline in three stages
(\textit{Search} $\rightarrow$ \textit{Filter} $\rightarrow$ \textit{Selection}).
The Naïve RAG baselines already reveal a clear mismatch between retrieving plausible candidates
and making correct multimodal decisions: although their overall \textit{Search} performance is
relatively reasonable (33.79/33.87 for Qwen3-8B/32B), the final \textit{Selection} quality is much
lower, largely because \textit{Image} selection is extremely weak (7.81 and 10.40), which further
drags down the overall selection scores to 27.03 and 28.07.
This gap suggests that the system can often retrieve plausible evidence, but struggles to
\emph{select} and \emph{insert} the correct images among the retrieved candidates, making image
selection the dominant bottleneck.

Building on this observation, our pipeline improves \emph{image search} in several settings (e.g.,
from 32.69 to 37.87 on Qwen3-8B, and up to 39.80 on Qwen3-8B$^\spadesuit$); however, better retrieval
alone does not necessarily yield better downstream \emph{image selection}.
In particular, without explicitly learning the insertion policy, the DPO variant remains weak on
image selection (e.g., 9.81 for Qwen3-8B$^\clubsuit$), whereas SFT substantially improves it (e.g.,
45.03 for Qwen3-8B$^\spadesuit$ and 41.39 for Qwen3-32B$^\spadesuit$), approaching GPT-4.1 (45.39).

This finding is consistent with the generation-quality results in Table~\ref{tab:deepreporter_generation}:
DPO yields mixed outcomes on text-centric dimensions (e.g., Qwen3-8B$^\clubsuit$ slightly improves
Fluency to 25.51 vs.\ 23.48 for Qwen3-8B), yet multimodal quality remains low (multimodal Avg 7.09 vs.\ 5.57)
and can even degrade overall performance in some settings (e.g., Qwen3-32B$^\clubsuit$ overall 18.35 vs.\ 27.16
for Qwen3-32B).
In contrast, SFT provides consistent gains on multimodal dimensions (multimodal Avg 41.09 for
Qwen3-8B$^\spadesuit$ and 41.70 for Qwen3-32B$^\spadesuit$) and achieves higher overall score
(29.01 and 37.89, respectively).

We attribute these trends to supervision granularity: multimodal insertion actions occupy only a
small fraction of the output context, and DPO relies on trajectory-level preference signals without
explicit, localized rewards for these sparse decisions, which makes it easier to optimize
coarse textual properties than fine-grained insertion behavior; in contrast, SFT optimizes every
token (including sparse multimodal-related tokens) via next-token prediction, providing a more
direct learning signal for image selection and placement.

\section{Qualitative Analysis of Agentic Multimodal Generation}
\label{app:case_studies}

We present three end-to-end examples illustrating the full agentic pipeline, including planning, multimodal search, relevance filtering, and incremental report generation.

\clearpage

\begin{center}
    \includegraphics[width=\textwidth]{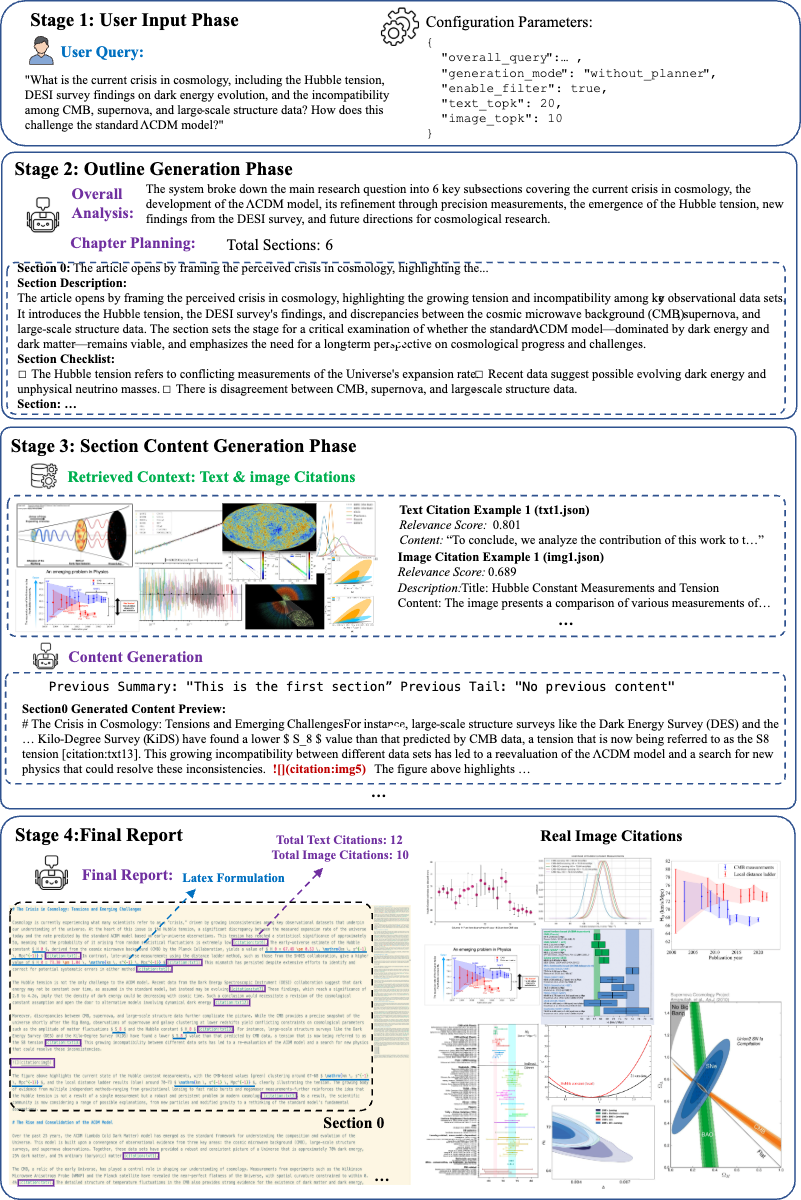}
\end{center}

\clearpage

\begin{center}
    \includegraphics[width=\textwidth]{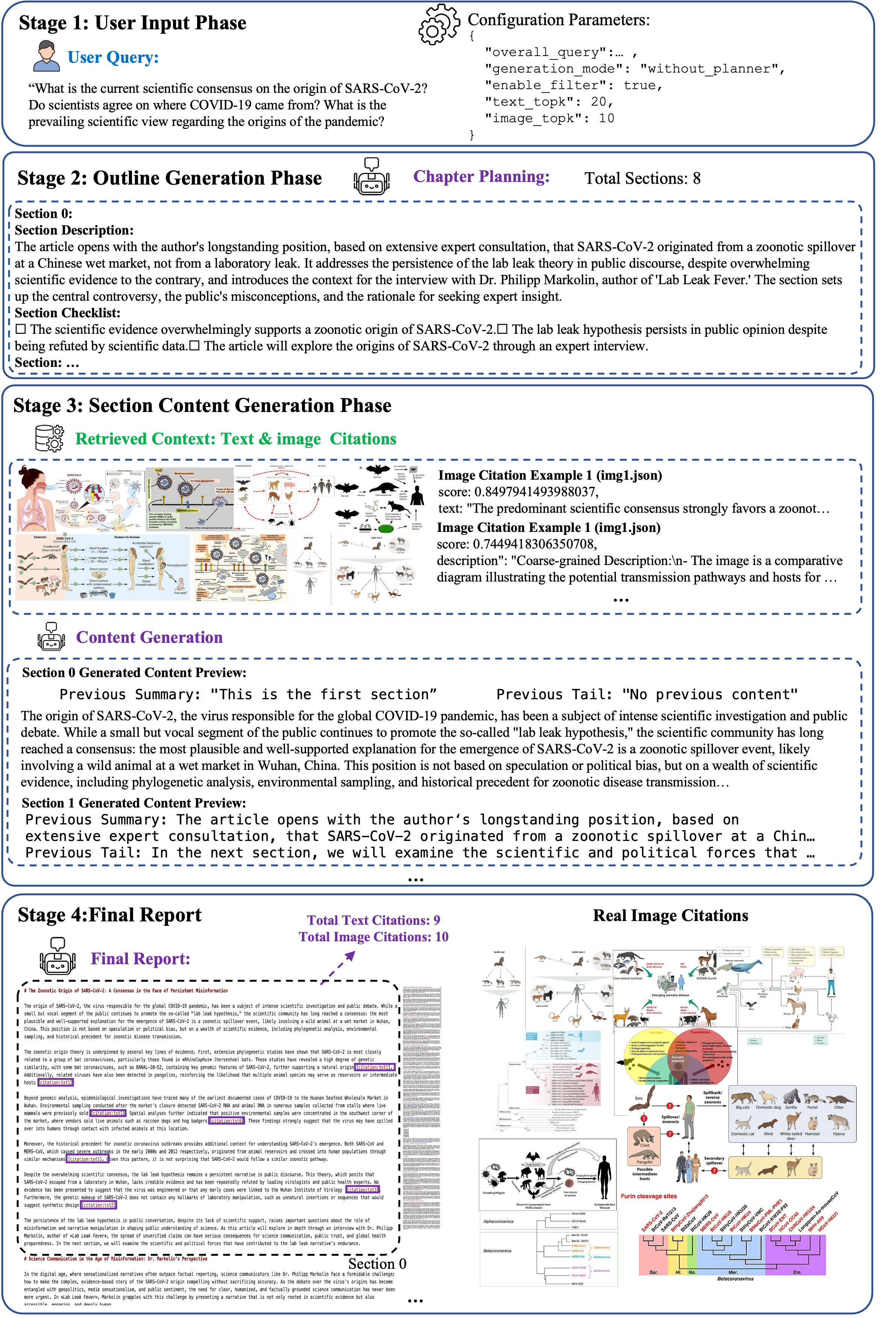}
\end{center}

\clearpage

\begin{center}
    \includegraphics[width=\textwidth]{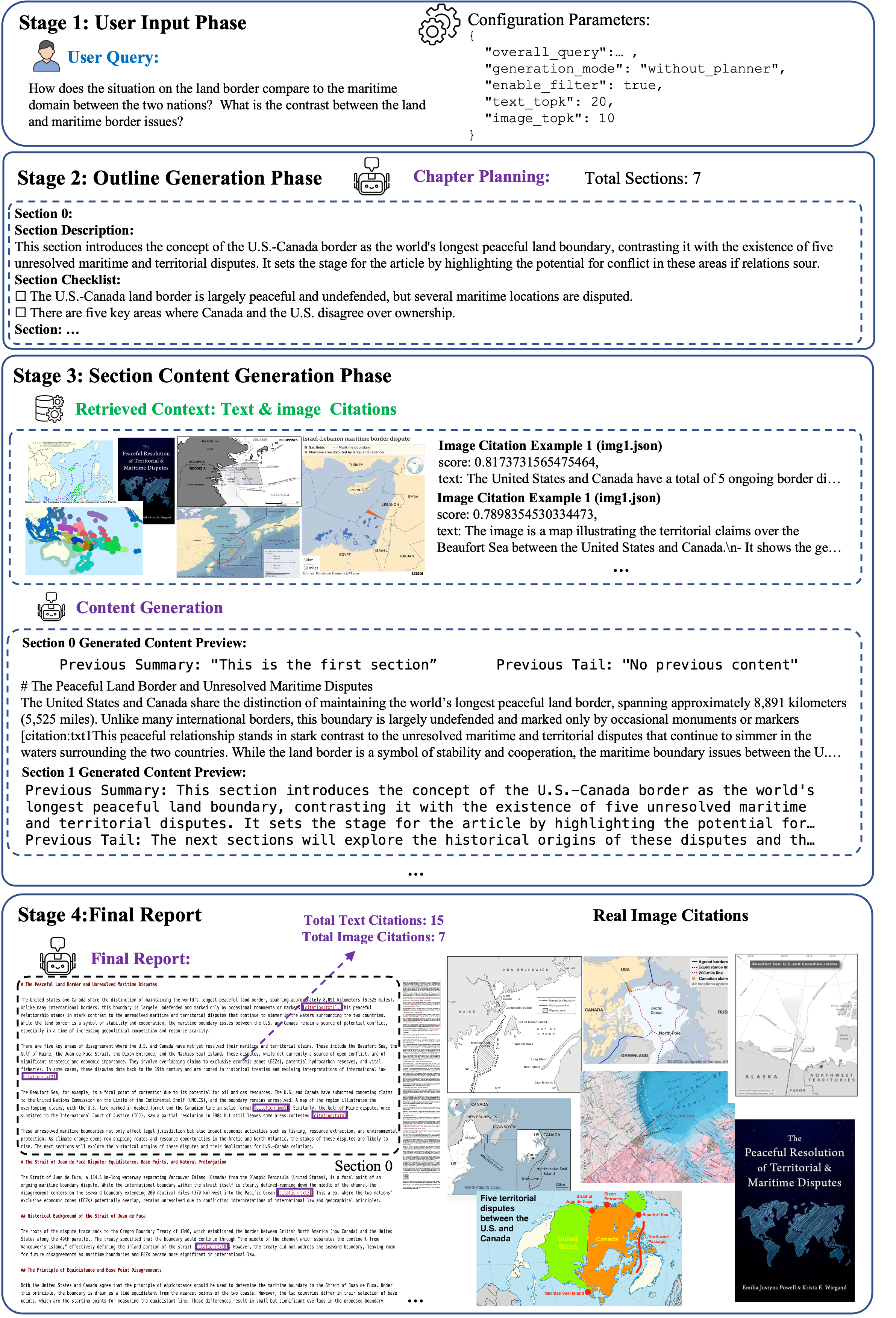}
\end{center}

\end{document}